\documentclass[a4paper, twoside, 12pt]{article}

\usepackage[noblocks]{authblk} 
\usepackage[numbers]{natbib}
\usepackage{graphicx}
\usepackage{fancyhdr}
\usepackage[colorlinks=true, linkcolor=MidnightBlue, urlcolor=MidnightBlue, citecolor=PineGreen]{hyperref}
\usepackage[tableposition=top, justification=justified, textfont=it]{caption}
\usepackage{titlesec}
\usepackage{multicol}
\usepackage{afterpage}
\usepackage{lipsum}
\usepackage[dvipsnames, x11names]{xcolor}
\usepackage[inner=4cm, outer=4cm, top=2.8cm, bottom=2.8cm, marginparsep=0cm, marginparwidth=0cm, includehead, includefoot]{geometry}
\usepackage{setspace}
\usepackage{array}
\usepackage{bm}
\usepackage{tikz}
\usepackage{float}
\usepackage[framemethod]{mdframed}
\usepackage{amsmath}
\usepackage{amssymb}

\titleformat{\section}{\Large \bfseries \centering \scshape}{\thesection.}{0.3em}{}[{\titlerule[0.5pt]}]

\definecolor{shadecolor}{RGB}{230,230,230}
\newcommand{\mybox}[1]{\par\noindent\colorbox{shadecolor}
{\parbox{\dimexpr\textwidth-2\fboxsep\relax}{#1}}}

\titleformat{\subsection}{\large \bfseries \mybox}{\thesubsection}{1em}{}

\titleformat{\subsubsection}{\itshape}{\thesubsubsection.}{0.3em}{}

\renewenvironment{abstract}
{\vskip 2.5ex {\large\bf\noindent Abstract}\vspace{0.7ex} \\ %
  \bgroup\noindent\ignorespaces}%
{\par\egroup\vskip 2.5ex}

\newenvironment{keywords}
{\bgroup\leftskip 20pt\rightskip 20pt \small\noindent{\bf Keywords:} }%
{\par\egroup\vskip 10ex}

\fancypagestyle{firstpage}{%
  
  \fancyhead[]{}
  \fancyfoot[C]{}
  \fancyfoot[L]{\footnotesize To appear in \\
  O. Colliot (Ed.), \textit{Machine Learning for Brain Disorders}, Springer\\
  }

}

\fancypagestyle{otherpages}{%
  
  \fancyhead[LE]{\thepage}
  \fancyhead[RE]{\runningauthor}
  \fancyhead[LO]{\runningheadtitle}
  \fancyhead[RO]{\thepage}
  \fancyfoot[L]{\footnotesize  \textit{Machine Learning for Brain Disorders}, Chapter~\chapternumber}
  \fancyfoot[C]{}
}

\makeatletter
\renewcommand{\maketitle}{\bgroup\setlength{\parindent}{0pt}

\begin{flushright}
  \color{MidnightBlue}
  \textbf{\LARGE Chapter~\chapternumber}
\end{flushright}

\vspace{0.3in}

\begin{flushleft}
    \setstretch{2.0} 
    \textbf{\color{MidnightBlue}\huge\@title}
\end{flushleft}

\vspace{0.15in}

\begin{flushleft}
    \textbf{\bfseries \large\@author}
\end{flushleft}\egroup
}

\makeatother

\DeclareCaptionLabelFormat{labelformat}{\color{MidnightBlue}\bfseries #1 #2}
\renewcommand{\bibpreamble}{\scriptsize \begin{multicols}{2}}
\renewcommand{\bibpostamble}{\end{multicols}}

\mdfsetup{skipabove=\topskip, skipbelow=\topskip}

\newcounter{nicebox}
\newenvironment{nicebox}[1][]{%
    \refstepcounter{nicebox}%
    \ifstrempty{#1}%
    {\mdfsetup{%
        frametitle={%
            \tikz[baseline=(current bounding box.east),outer sep=0pt]
            \node[anchor=east,rectangle,fill=MidnightBlue]
            {\strut Theorem~\thetheo};}}
    }%
    {\mdfsetup{%
        frametitle={%
            \tikz[baseline=(current bounding box.east),outer sep=0pt]
            \node[anchor=east,rectangle,fill=MidnightBlue]
            {\strut \color{white} Box~\thenicebox:~#1};}}%
    }%
    \mdfsetup{innertopmargin=10pt,linecolor=MidnightBlue,     
    linewidth=2pt,topline=true, frametitleaboveskip=\dimexpr-\ht\strutbox\relax,}
    \begin{mdframed}[]\relax%
    }{\end{mdframed}}

\newfloat{floatbox}{thp}{lop}
\floatname{floatbox}{Box}

\DeclareMathAlphabet{\mathsfit}{\encodingdefault}{\sfdefault}{m}{sl}
\SetMathAlphabet{\mathsfit}{bold}{\encodingdefault}{\sfdefault}{bx}{n}

\usepackage{caption}
\usepackage{subcaption}

\begin{document}


\newcommand{\runningauthor}{Wenzel}

\newcommand{\runningheadtitle}{GANs and Beyond}

\newcommand{\chapternumber}{5}

\newcommand{\emailaddress}{markus.wenzel@mevis.fraunhofer.de}

\title{Generative Adversarial Networks and other Generative Models} 

\author[*,1]{Markus Wenzel}  

\affil[1]{Fraunhofer Institute for Digital Medicine MEVIS, Bremen, Germany}

\affil[*]{Corresponding author: e-mail address: \href{mailto:\emailaddress}{\emailaddress}}

\maketitle

\afterpage{\aftergroup\restoregeometry}
\pagestyle{otherpages}

\begin{abstract}
Generative networks are fundamentally different in their aim and methods compared to CNNs for classification, segmentation, or object detection. They have initially not been meant to be an image analysis tool, but to produce naturally looking images. The adversarial training paradigm has been proposed to stabilize generative methods, and has proven to be highly successful -- though by no means from the first attempt.

This chapter gives a basic introduction into the motivation for Generative Adversarial Networks (GANs) and traces the path of their success by abstracting the basic task and working mechanism, and deriving the difficulty of early practical approaches. Methods for a more stable training will be shown, and also typical signs for poor convergence and their reasons. 

Though this chapter focuses on GANs that are meant for image generation and image analysis, the adversarial training paradigm itself is not specific to images, and also generalizes to tasks in image analysis. Examples of architectures for image semantic segmentation and abnormality detection will be acclaimed, before contrasting GANs with further generative modeling approaches lately entering the scene. This will allow a contextualized view on the limits but also benefits of GANs.   
\end{abstract}

\begin{keywords}
generative models, generative adversarial networks, GAN, CycleGAN, StyleGAN, VQGAN, diffusion models, deep learning
\end{keywords}

\section{Introduction}
\label{sec:intro}

\begin{figure}[hbtp]
	\centering
		\includegraphics[width=0.85\textwidth]{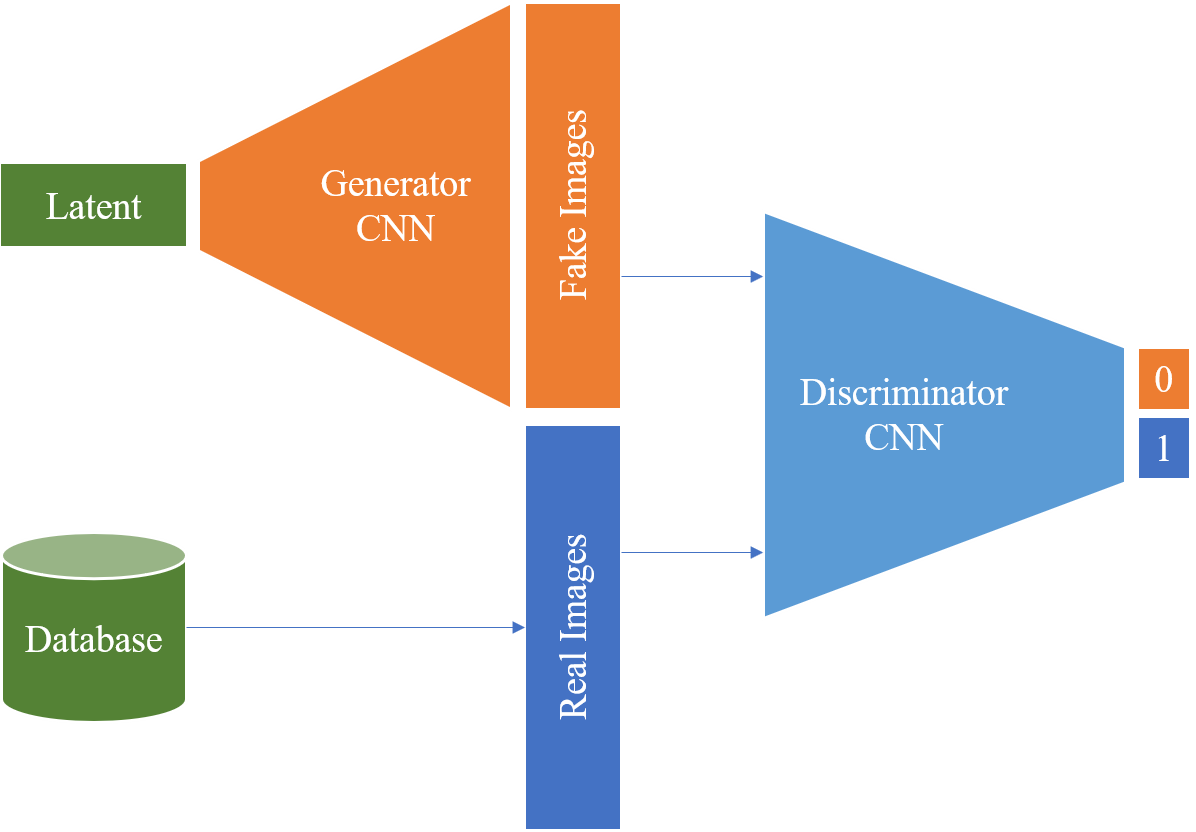}
	   \caption[GAN Approach]{The fundamental GAN setup for image generation consisting of a Generator and a Discriminator network; here: CNNs.}
	\label{fig:GAN-Intro}
\end{figure}

Generative Adversarial Networks are a type of neural network architecture, in which one network part generates solutions to a task, and another part compares and rates the generated solutions against a priori known solutions. While at first glimpse this does not sound much different from any loss function, which essentially also compares a generated solution with the gold standard, there is one fundamental difference. A loss function is static, but the ``judge'' or ``discriminator'' network part is trainable (\autoref{fig:GAN-Intro}). This means that it can be trained to distinguish the generated from the true solutions and, as long as it succeeds in its task, a training signal for the generative part can be derived. This is how the notion of adversaries came into the name GAN. The discriminator part is trained to distinguish true from generated solutions, while the generative part is trained to arrive at the most realistic appearing solutions, making them adversaries with regard to their aims.

Generative Adversarial Networks are now among the most powerful tools to create naturally looking images from many domains. While they have been created in the context of image generation, the original publication describes the general idea of how to make two networks learn by competing, regardless of the application domain. This key idea can be applied to generative tasks beyond image creation, including text generation, music generation, and many more.

The research interest skyrocketed in the years after the first publication proposing an adversarial training paradigm~\cite{Goodfellow2014GAN}. Looking at the number of web searches for the topic ``generative adversarial networks'' shows how the interest in the topic has rapidly grown, but also the starting decline of the last years. Authors since 2014 have cast all kinds of problems into the GAN framework, to enable this powerful training mechanism for a variety of tasks, including image analysis tasks as well. This is surprising at first, since there is no immediate similarity between a generative task and, for example, a segmentation or detection task. Still, as evidenced by the success in these application areas, the adversarial training approach can be applied with benefits. Clearly, the decline in interest can to some degree be attributed to the emergence of best practices and proven implementations, while simultaneously the scientific interest has recently shifted to successor approaches. However, similar to the persistent relevance of CNN architectures like ResNets for classification, Mask R-CNNs for detection, or basic Transformer architectures for sequence processing, GANs will remain an important tool for image creation and image analysis. The adversarial training paradigm has become an ingredient to models apart from generative aims, providing flexible ways to custom-tailor loss components for given tasks.

\begin{figure}[hbtp]
	\centering
		\includegraphics[width=0.95\textwidth]{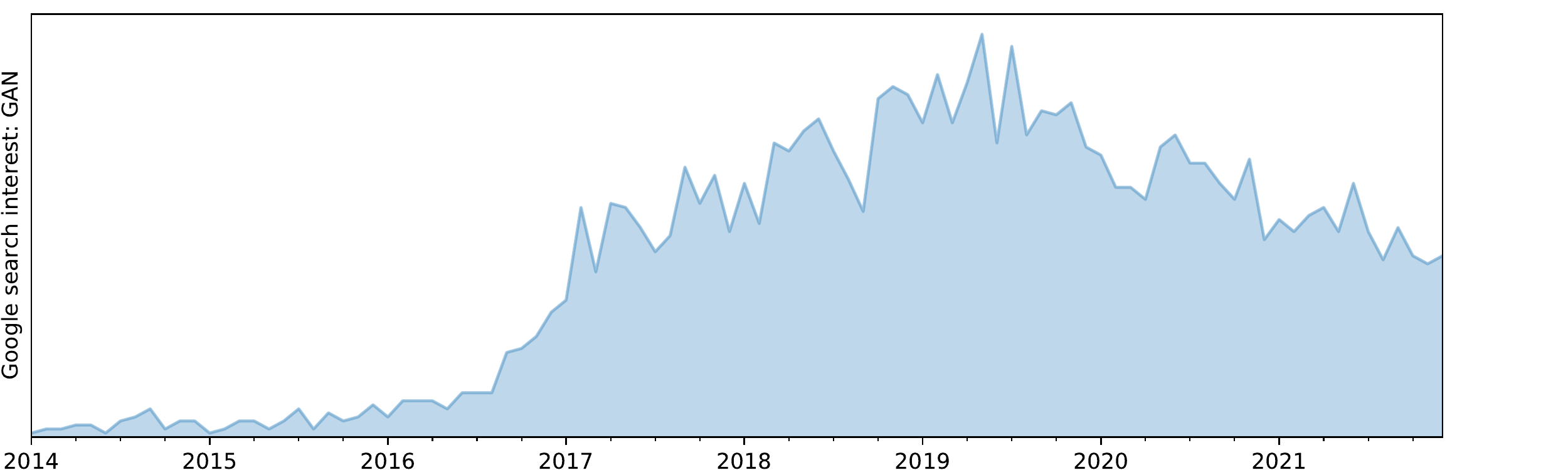}
	   \caption[GAN search trends]{Google web search based interest estimate for ``generative adversarial networks'' since 2014. Relative scale.}
	\label{fig:searchtrend}
\end{figure}

\begin{figure}[hbtp]
	\centering
		\includegraphics[width=0.95\textwidth]{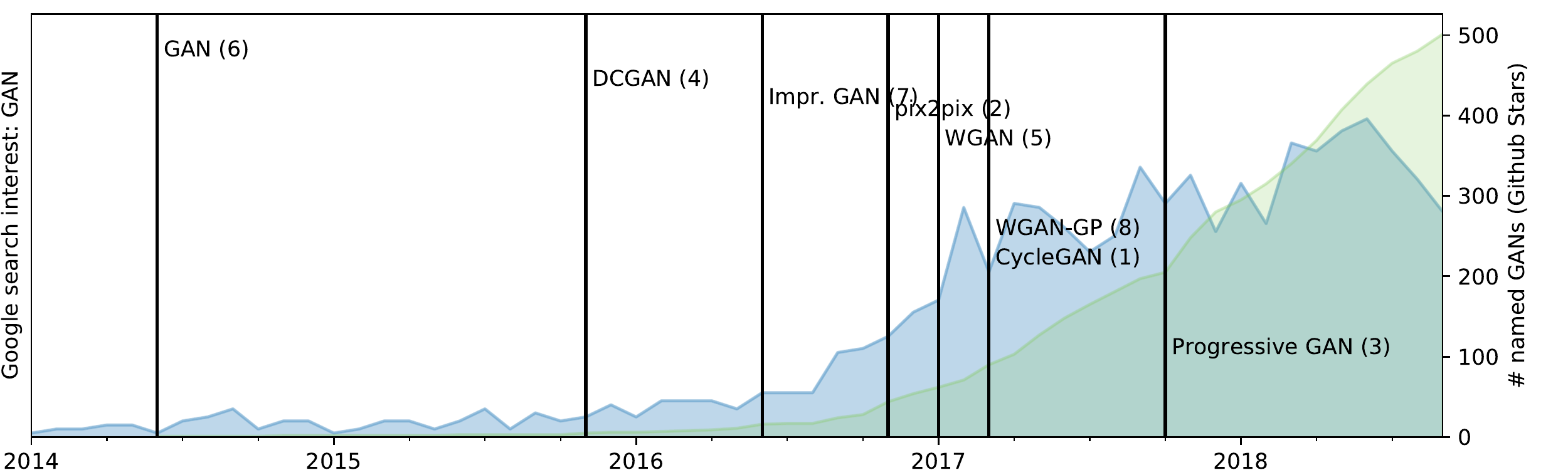}
	   \caption[GAN names]{Some of the most-starred shared GAN code repositories on Github, until 2018. Ranking within this selection in brackets.}
	\label{fig:gannames}
\end{figure}

\section{Generative Models} 
\label{sec:Gen_Models}

Generative processes are fundamentally hard to grasp computationally. Their nature and purpose is to create something ``meaningful'' out of something less meaningful (even random). The first question to ask therefore is how this can even be possible for a computer program since, intuitively, creation requires an inventive spirit -- call it creativity, to use the term humans tend to associate with this. To introduce some of the terminology and basic concepts that we will use in the remainder of this section, some remarks on human creativity will set the scene. 

In fact, creative human acts are inherently limited by our concepts of the world, acquired by learning and experience through the sensory means we have available, and by the available expressive means (tools, instruments, ...) with which we can even conceive of creating something. This is true for any kind of creative act, including writing, painting, wood carving or any other art, and similarly also for computer programming, algorithm development, or science in general. Our limited internal representation of the world around us frames our creative scope. 

This is very comparable to the way computerized, programmed or learned generative processes create output. They have either an in-build mechanism, or a way to acquire such a mechanism, that represents the tools by which creation is possible, as well as a model of the world that defines the scope of outputs. Practically, a CNN-based generative process uses convolutions as the in-built tool, and is by this tool geared to produce image-like outputs. The convolutional layers, if not a priori defined, will represent a set of operations defined by a training process and limited in their expressiveness by the training material -- by the fraction of the world that was presented. This will lead us to the fundamental notion of how to capture the variability of the ``fraction of the world'' that is interesting, and how to make a neural network represent this partial world knowledge. It is interesting to note at this point, that neither for human creative artists nor for neural networks the ability to (re-)create convincing results implies an understanding of the way, the templates (in the real world) have come into existence. Generating convincing artifacts does not imply understanding nature. Therefore, GANs cannot explain the parts of nature they are able to generate.

\subsection{The Language of Generative Models: Distributions, Density Estimation, Estimators}
\label{sec:gan_language}

Understanding the principles of generative models requires a basic knowledge of distributions. The reason is that -- as already hinted at in the previous section -- the ``fraction of the world'' is in fact something that can be thought of as a distribution in a parameter space. If you were to describe a part of the world in a computer-interpretable way, you would define descriptive parameters. To describe persons, you could characterize them by simple measures like age, height, weight, hair and eye color, and many more. You could add blood pressure, heart rate, muscle mass, maximum strength and more, and even a whole-genome sequencing result might be a parameter. Each of the parameters individually can be collected for the world population, and you will obtain a picture of how this parameter is ``distributed'' world-wide. In addition, parameters will be in relation with each other, for example age and maximum strength. Countless such relationships exist, of which the majority are and probably will remain unknown. Those interrelationships are called a joint distribution. Would you know the joint distribution, you could ``create'' a plausible parameter combination of a non-existing human. Let us formalize these thoughts now.

\subsubsection{Distributions}
\label{sec:distributions}

A distribution describes the frequency of particular observations when watching a random process. Plotting the number of occurrences over an axis of all possible observations creates a histogram. If the possible observations can be arranged on a continuous scale, one can see that observations cluster in certain areas, and we say that they create a ``density'', or are ``dense'' there. Hence, when trying to describe where densities are in parameter space, this is associated with the desire to reproduce or sample from distributions, like we want to do it to generate instances from a domain. Before being able to reproduce the function that generates observations, estimating where the dense areas are is required. This will in the most general sense be called density estimation. 

Sometimes, the shape of the distribution follows an analytical formula, for example the Normal distribution. If such a closed form description of the distribution can be given, like for instance the Normal distribution, this distribution generalizes the shape of the histogram of observations and makes it possible to produce new observations very easily, by simply sampling from the distribution. When our observations follow a Normal distribution, we mean that we expect to observe instances more frequently around the mean of the Normal distribution than towards the tails. In addition, the standard deviation quantifies how much more likely observations close to the mean are compared to observations in the tails. We describe our observations with a parametric description of the observed density.

In the remainder of this section, rather than providing a rigorous mathematical definition and description of the mathematics of distributions and (probability) density estimation, we will introduce the basic concepts and terminology in an intuitive way (also compare \autoref{box:distrib}). Readers with the wish for a more in-depth treatment can find tutoring material in the references~\cite{casella2021statistical,grinstead2006introduction,severini2005elements,pml0Book,pml1Book}.

\begin{floatbox}[htbp]
    \begin{nicebox}[Probability Distributions -- Terminology]
        \label{box:distrib}
        Several common terms regarding distributions have intuitive interpretations which are given in the following. Let $a$ be an event from the probability distribution $A$, written as $a \sim A$, and $b\sim B$ an event from another probability distribution. 
        
        In a medical example, $A$ might be the distribution of possible neurological diseases, and $B$ the distribution of all possible variations of smoking behaviour.
        
        \begin{description}
            \item[Conditional Probability $P(A \vert B)$] The conditional probability of a certain $a \sim A$, for example a stroke, might depend on the concrete smoking history of a person, described by $b\sim B$. The conditional probability is written as $p(a \vert b)$ for the concrete instances, or $P(A \vert B)$ if talking about the entire probability distributions $A$ and $B$. 
            \item[Joint Probability $P(A,B)$] The probability of seeing instantiations of $A$ and $B$ together is termed the joint probability. Notably, if expanded, this will lead to a large table of probabilities, joining each possible $a \sim A$ (for example stroke, dementia, Parkinson's disease, etc.) with each possible $b \sim B$ (Casual smoker, frequent smoker, non-smoker, etc.). 
            \item[Marginal Probability] The marginal probabilities of $A$ and $B$ (denoted respectively P(A) and P(B)) are the probabilities of each possible outcome across (and independent of) all of the possible outcomes of the other distribution. For example, it is the probability of seeing non-smokers across all neurological diseases, or seeing a specific disease regardless of smoking status. It is said to be the probability of one distribution marginalized over the other probability distributions.
        \end{description}
    \end{nicebox}
\end{floatbox}

\subsubsection{Density estimation}
\label{sssec:density}
We assume in the following that our observations have been produced by a function or process that is not known to us, and that cannot be guessed from an arrangement of the observations. In a practical example, the images from a CT or MRI scanner are produced by such a function. Notably, the concern is less about  the intractability of the imaging physics, but about the appearance of the human body. The imaging physics might be modeled analytically up to a certain error. But the outer shape and inner structure of the human body and its organs depend on a large amount of mutually influencing factors. Some of these factors are known and can even be modeled, but many are not. In particular the interdependence of factors must be assumed to be intractable. What we can accumulate is measured data providing information about the body, its shape, and its function. While many measurement instruments exist in medicine, for this chapter, we will be concerned with images as our observations. In the following thought experiment we will explore a na\"{i}ve way to model the distribution and try to generate images. 

\begin{figure}[hbt]
	\centering
		\includegraphics[width=0.34\textwidth]{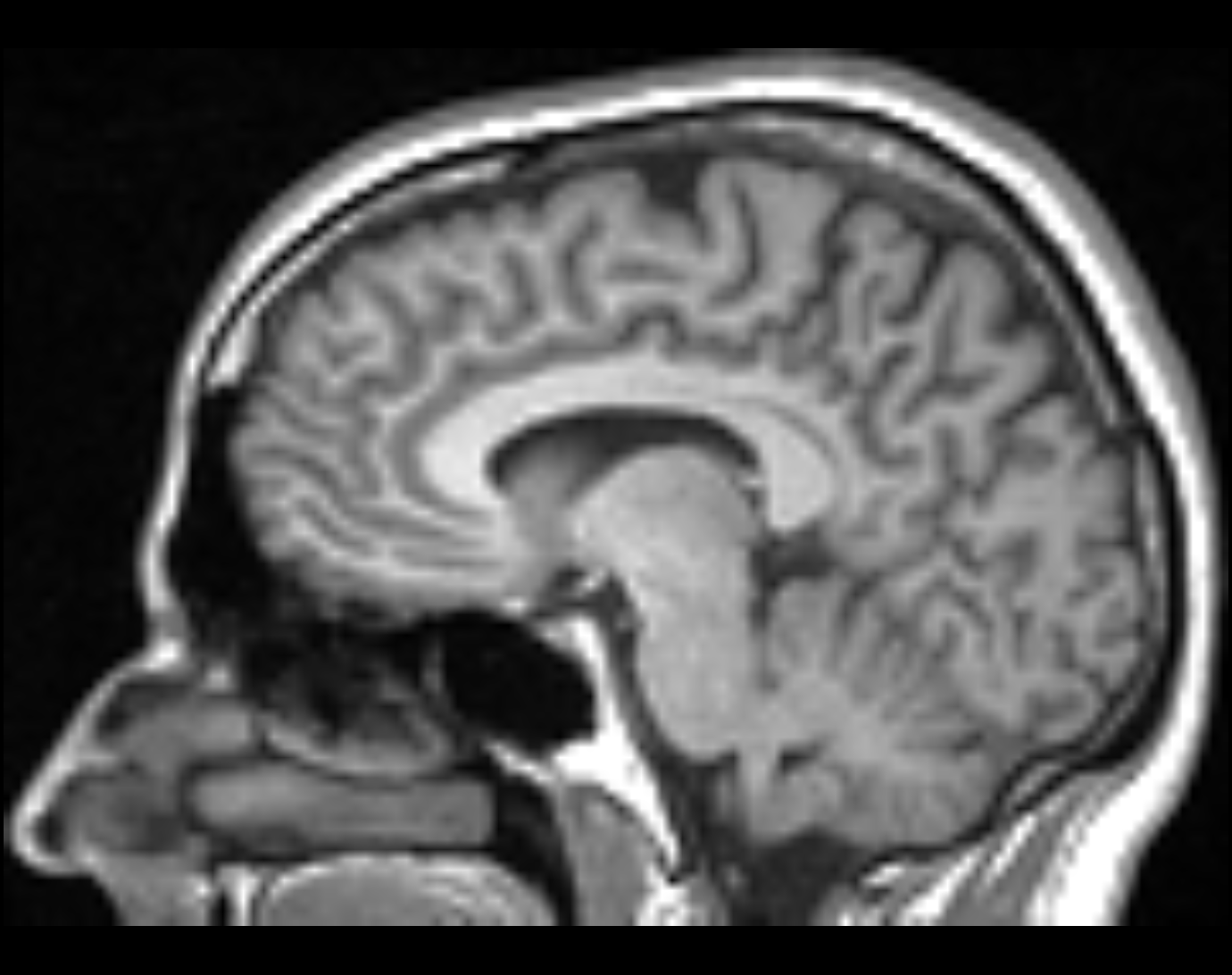}
		\includegraphics[width=0.63\textwidth]{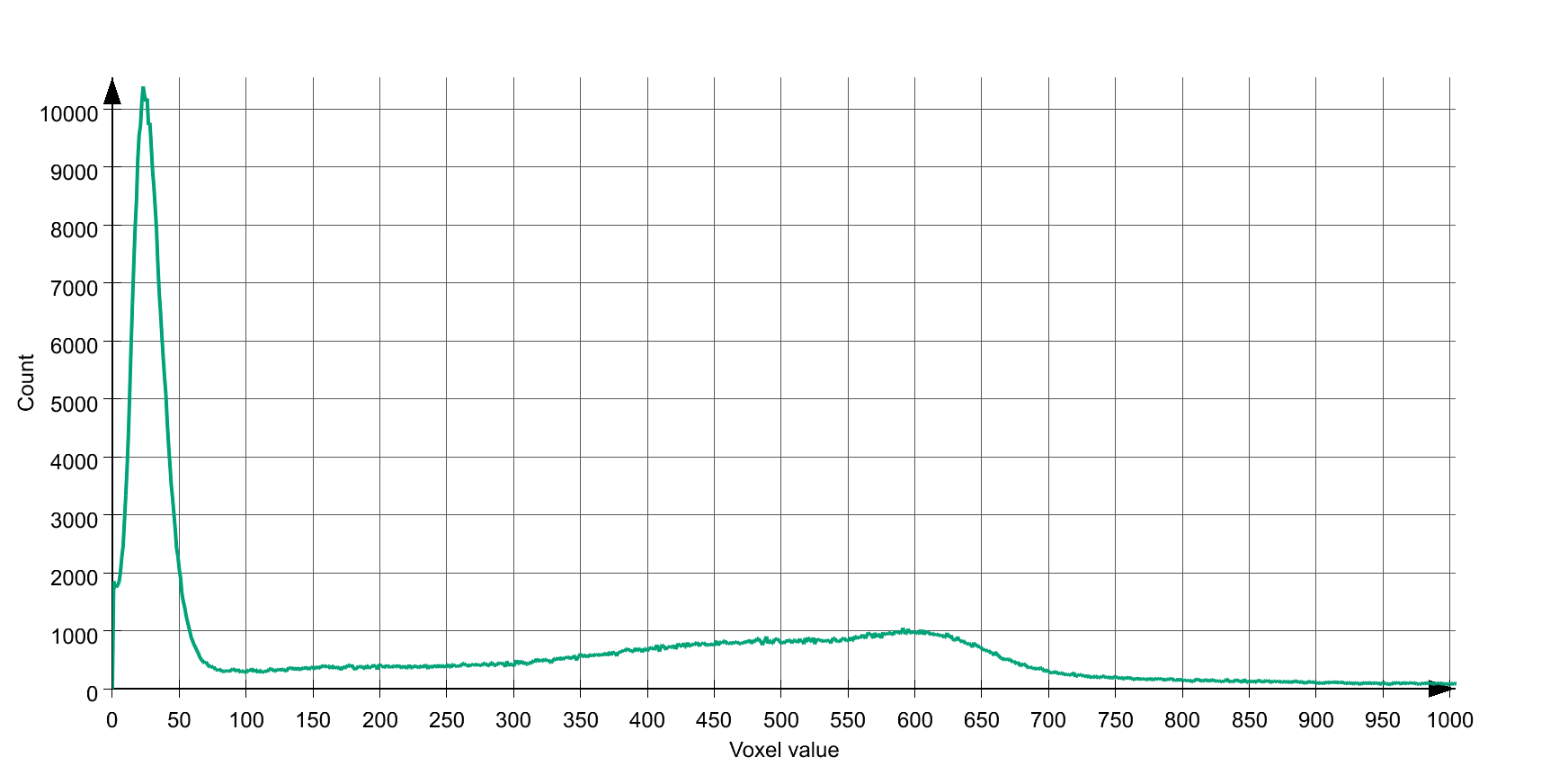}
	   \caption[MRI Histogram]{Brain MRI (left) and histogram of gray values for one slice of a Brain MRI.}
	\label{fig:mri-histo}
\end{figure}

\begin{figure}[hbt]
	\centering
		\includegraphics[width=0.34\textwidth]{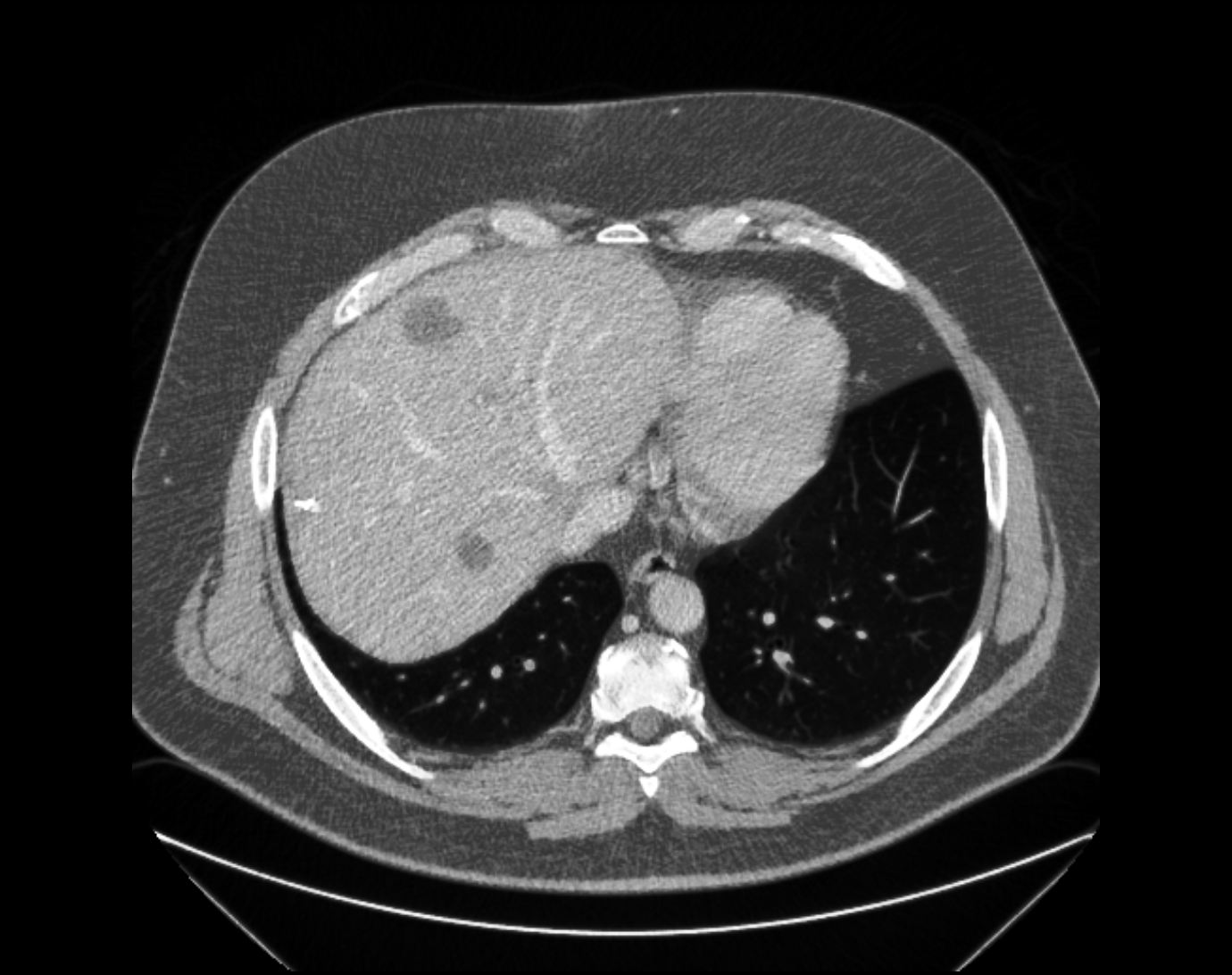}
		\includegraphics[width=0.63\textwidth]{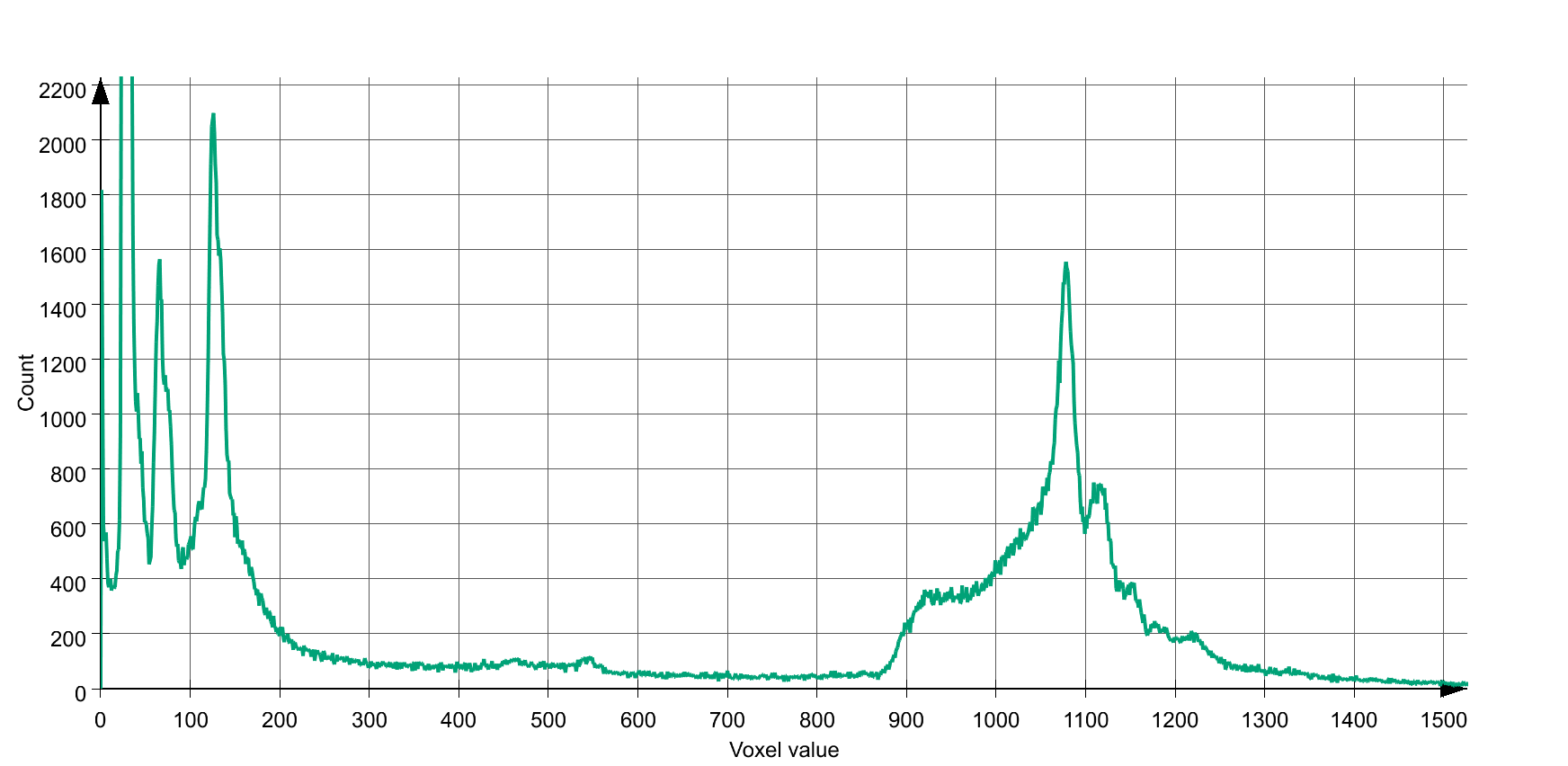}
	   \caption[CT Histogram]{Abdominal CT (left) and histogram of gray values for one slice of an abdominal CT}
	\label{fig:ct-histo}
\end{figure}

The first step is to examine the gray value distribution, or in other words, estimate the density of values. The most basic way for estimating a density is plotting a histogram. Let the value on the x axis be the image gray value of the medical image in question (in CT expressed in Hounsfield units, HU, and in arbitrary units for MRI). Two plots show histograms of a head MRI (\autoref{fig:mri-histo}) and an abdominal CT (\autoref{fig:ct-histo}). While the brain MRI suggests three or four major ``bumps'' of the histogram at about values 25, 450, and 600, the abdominal CT doesn't lend itself to such a description.

In the next step, we want to describe the histograms through analytical functions, to make them amenable for computational ends. This means, we will aim to estimate an analytical description of the observations.

Expectation Maximisation (EM, see \autoref{box:em_example}) is an algorithm suitable for this task. EM enables us to perform maximum likelihood estimation in the presence of unobserved (``latent) variables and incomplete data -- this being the default assumption when dealing with real data. Maximum likelihood estimation (MLE) is the process of finding parameters of a parametric distribution to most accurately match the distribution to the observations. In MLE, this is achieved by adapting the parameters steered by an error metric that indicates the closeness of the fit; in short, a parameter optimization algorithm. 

\begin{floatbox}[htbp]
    \begin{nicebox}[Expectation Maximisation -- Example]
        \label{box:em_example}

        Focusing on our density estimate of the MRI data, we want to use Expectation Maximisation (EM) to optimize the parameters of a fixed number of Gaussian functions adding up to the closest possible fit to the empirical shape of the histogram.
        
        In our data, we observe ``bumps'' of the histogram. We can by image analysis determine that certain organs imaged by MRI lead to certain bumps in the histogram, since they are of different material and create different signal intensities. This, however, is unknown to EM -- the so-called ``latent'' variables.
        
        The EM algorithm has two parts, the Expectation step and the Maximisation step. They can, with quite far-reaching omission of details, be sketched as follows.
        
        \begin{description}
            \item[Expectation] takes each point (or a number of sampled points) of the distribution and \emph{estimates the expectation} to which of the parameterized distribution to assign it to. Figuring out this assignment is the step of dealing with the ``latent'' variable of the observations.
            \item[Maximisation] iterates over all parameterized distributions and adjusts their parameters to match the assigned points as well as possible. 
        \end{description}

        This process is iterated until a fitting error cannot be improved anymore.
        
        An short introductory treatment of EM with examples and applications is presented in~\cite{Do2008EM}. The standard reference for the algorithm is~\cite{Dempster1977EM}.
    \end{nicebox}
\end{floatbox}

\begin{figure}[hbt]
	\centering
		\includegraphics[width=0.93\textwidth]{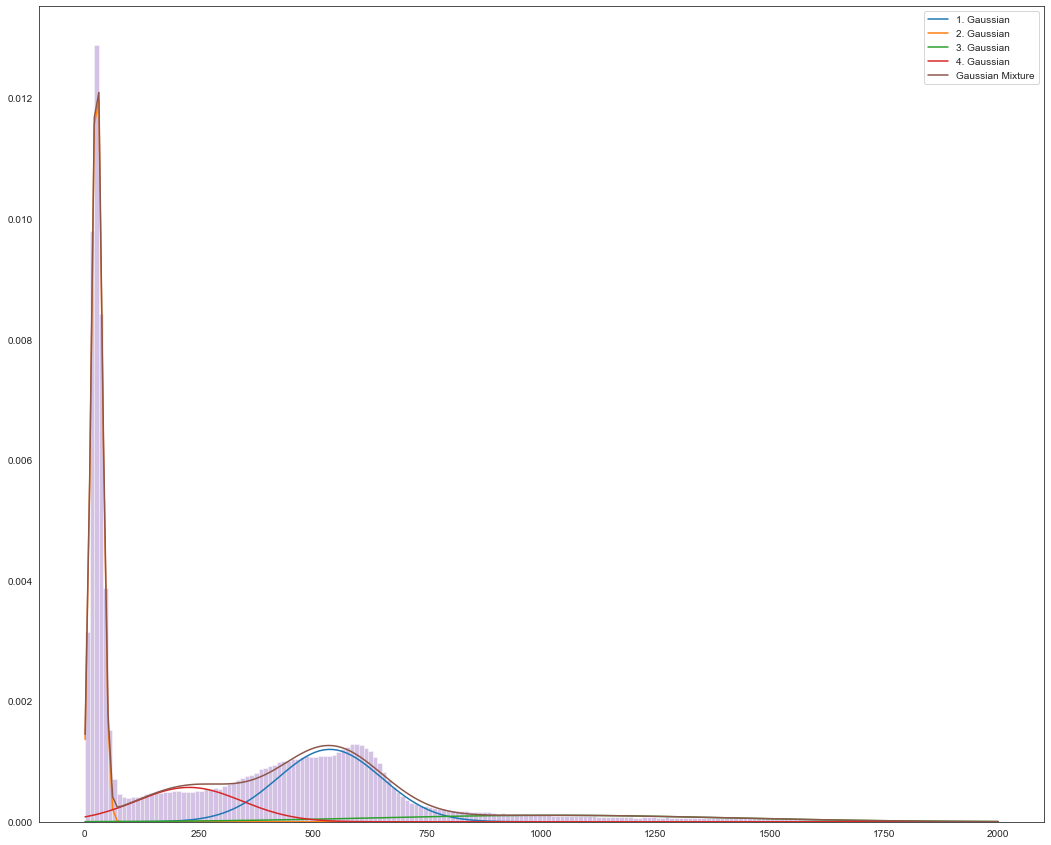}
	   \caption[EM Fit Example]{A Gaussian Mixture Model (GMM) of four Gaussians was fit to the Brain MRI data we have visualized as a histogram in \autoref{fig:mri-histo}.}
	\label{fig:em-on-brain-mri}
\end{figure}

In \autoref{fig:em-on-brain-mri}, a mixture of four Gaussian distributions has been fit to the brain MRI voxel value data seen before. 

It is tempting to model even more complex observations by mixing simple analytical distributions (e.g. Gaussian Mixture Models, GMMs), but in general this will be intractable for two reasons. Firstly, realistic joint distributions will have an abundance of mixed maxima and therefore require a vast number of basic distributions to fit. Even basic normal distributions in high-dimensional parameter spaces are no longer functions with two parameters $(\mu,\sigma)$, but with a vector of means and a covariance matrix. Secondly, it is no longer trivial to sample from such high-dimensional joint distributions, and while some methods, among others Markov Chain Monte Carlo methods, allow to sample from them, such numerical approaches are of such high computational complexity that it makes their use difficult in the context of deep neural network parameter estimation.

We will learn about alternatives. In principle, there are different approaches for density (distribution) estimation: direct distribution estimation, distribution approximation, or even more indirectly, by using a simple surrogate distribution that is made to resemble the unknown distribution as good as possible through a mapping function. We will see this in the further elaboration of generative modeling approaches.

\subsubsection{Estimators and the Expected Value}%
\label{sssec:estimators}

Assume we have found suitable mean values and standard deviations for three Normal distributions that together approximate the shape of the MRI data density estimate to our satisfaction. Such a combination of Normal (Gaussian) distributions is called a Gaussian Mixture Model or GMM, and sampling from such a GMM is straight-forward. We are thus able to sample single pixels in any number, and over time we will sample them such that their density estimate or histogram will look similar to the one we started with. 

However, if we want to generate a brain MRI image using a sampling process from our closed-form GMM representation of the distribution, we will notice that a very important notion wasn't respected in our approach. We start with one slice of $512 \times 512$ voxels, and therefore randomly draw the required number of voxel values from the distribution. However, this will not yield an image that resembles one slice of a brain MRI, but will almost look like random noise, because we did not model the spatial relation of the gray values with respect to each other. Since the majority of voxels of a brain MRI are not independent of each other, drawing one new voxel from the distribution needs to depend on the spatial locations and gray values of all voxels drawn before. Neighboring voxels will have a higher likelihood of similar gray values than voxels far apart from each other, for example. More crucially, underneath the interdependence lies the image generation process: the image values observed in a real brain MRI stem from actual tissue – and this is what defines their interdependence. This means, the anatomy of the brain indirectly reflects itself in the rules describing the dependency of gray values of one another.

For the modeling process, this implies that we cannot argue about single voxel values and their likelihood, but we need to approach the generative process differently. One idea for a generative process has been implied in the above description already: pick a random location of the to-be-generated image, and predict the gray value depending on all existing voxel values. Implemented with the method of mixture models, this results in unfathomably many distributions to be estimated, as for each possible ``next voxel'' location, any possible combination of already existing voxel numbers and positions needs to be considered. We will see in \autoref{sec:diffusion} on Diffusion Models how this general approach to image generation can still be made to work.

A different sequential approach to image generation has also been attempted, in which pixels are generated in a defined order, starting at the top left, and scanning the image row by row across the columns. Again, the knowledge about the already produced pixels is memorized and used to predict the next voxel. This has been dubbed the PixelRNN (Pixel Recurrent Neural Network), which lends its general idea from text processing networks~\cite{Oord2016PixelRN}.

Lastly, a direct approach to image generation could be formulated by representing or approximating the full joint distribution of all voxels in one distribution that is tangible, and to sample all voxels \emph{at once} from this. The full joint distribution in this approach remains implicit, and we use a surrogate. This will actually be the approach implemented in GANs, though not in a na\"{i}ve way.

Running the numbers of what a likelihood-based na\"{i}ve approach implies, the difficulties of making it work will become obvious. Consider an MRI image as the joint distribution of $512 \times 512$ voxels (one slice of our brain MRI), where we approximated the gray value distribution of one voxel with a GMM with six parameters. This results in a joint distribution of $512 \times 512 \times 6=1,572,864$ parameters. Conceptually, this representation therefore spans a 1,572,864-dimensional space, in which every one brain MRI slice will be one data point. Referring back to the histograms of CT and MRI images in the Figures above, we have seen continuous lines with densities because we have collected all voxels of an entire medical image, which are many million. Still, we only covered one single dimension out of the roughly 1.5 million. Searching for the density in the 1,572,864-dimensional MRI-slice-space that is given by all collected brain MRI slices is the difficult task any generative algorithm has to solve. In this vastly large space, the brain MRI slices ``live'' in a very tiny region that is extremely hard to find. We say the images occupy a low-dimensional manifold within the high-dimensional space. 

Consider the Maximum Likelihood formulation

\begin{equation}
    \label{eqn:max-likelihood}
    \hat{\theta} = \arg\max_\theta \mathbb{E}_{x \sim P_\mathrm{data}} \log Q_{\theta}(x|\theta)
\end{equation}

where $P_\mathrm{data}$ is the unknown data distribution, $Q_\theta$ the distribution generated by the model which is parameterized by $\theta$. $\theta$ can for example be the weights and biases of a deep neural network\footnote{We will use $\theta$ when referring to parameters of models in general, but designate parameters of neural networks with $w$ in accordance with literature.}. In other words, the result of maximum likelihood estimation are parameters $\hat{\theta}$ so that the product of two terms, out of which only the second depends on the choice of $\theta$, is maximal. The first term is the expectation of $x$ with regard to the real data distribution. The second term is the (log of) the conditional probability (likelihood) of seeing the example $x$ given the choice of $\theta$ under the model $Q_\theta$. Hence, maximizing the likelihood function means maximizing the probability that $x$ is seen in $Q_\theta$, which will be the case when $Q$ matches $P$ as closely as possible given the parametric form of $Q$.

The maximum likelihood mechanism is very nicely illustrated in~\cite{magnussonLikelihood}. Here, it is also visually shown how finding the maximum likelihood estimate of parameters of the distribution can be done by working with partial derivatives of the likelihood function with respect to $\mu$ and $\sigma^2$ and seeking their extrema. The partial derivatives are called the score function and will make a reappearance when we discuss score-based and diffusion models later in \autoref{sec:diffusion} on advanced generative models.

\subsubsection{Sampling from Distributions}
\label{sec:sampling}
When a distribution is a model of how observed values occur, then sampling from this distribution is the process of generating random new values that could have been observed, with a probability similar to the probability to observe this value in reality. There are two basic approaches to sampling from distributions: generating a random number from the uniform distribution (this is what a random number generator is always doing underneath), and feeding this number through the inverse Cumulative Density Function (iCDF) of the distribution, which is the function that integrates the probability density function (PDF) of the distribution. This can only be achieved if the CDF is given in closed form. If it is not, the second approach to sampling can be used, which is called acceptance (or rejection) sampling. With $f$ being the PDF, two random numbers $x$ and $y$ are drawn from the uniform distribution. The random $x$ is accepted, if $f(x)>y$, and rejected otherwise.

Our use case, as we have seen, involves not only high-dimensional (multivariate) distributions, but even more, their joints, and they are not given in closed form. In such scenarios, sampling can be done still, using Markov Chain Monte Carlo (MCMC) sampling, which is a framework using rejection sampling with added mechanisms to increase efficiency. While MCMC has favourable theoretic properties, it is still computationally very demanding for complex joint distributions, which leads to important difficulties in the context of sampling from distributions we are facing in the domain of image analysis and generation.

We are therefore at this point facing two problems: we can hardly hope to be able to estimate the density, and even if we could, we could practically not sample from it.

\section{Generative Adversarial Networks}
\label{sec:GANs}

\subsection{Generative vs. Discriminative Models}
\label{ssec:gen_vs_disc}

To emphasize the difficulty that generative models are facing, compare them to discriminative models. Discriminative models solve tasks like classification, detection, and segmentation, to name some of the most prominent examples. How classification models are in the class of discriminative models is obvious: discriminating examples is exactly classifying them. Detection models are also discriminative models, though in a broader sense, in that they classify the detection proposals into accepted object detections or rejected proposals, and even the bounding box estimation, which is often solved through bounding box regression, typically involves the discriminative prediction of template boxes. Segmentation, on the other hand, for example using a U-Net, is only the extension of classic discriminative approaches into a fast framework that avoids pixel-wise inference through the model. It is common to all these models that they yield output corresponding to their input, in the sense that they extract information from the input image (for example, an organ segmentation, a classification, or even a textual description of the image content) or infer additional knowledge about it (for example a volume measurement, or an assessment or prediction of a treatment success given the appearance of the image).

\begin{figure}[hbtp]
	\centering
		\includegraphics[width=0.85\textwidth]{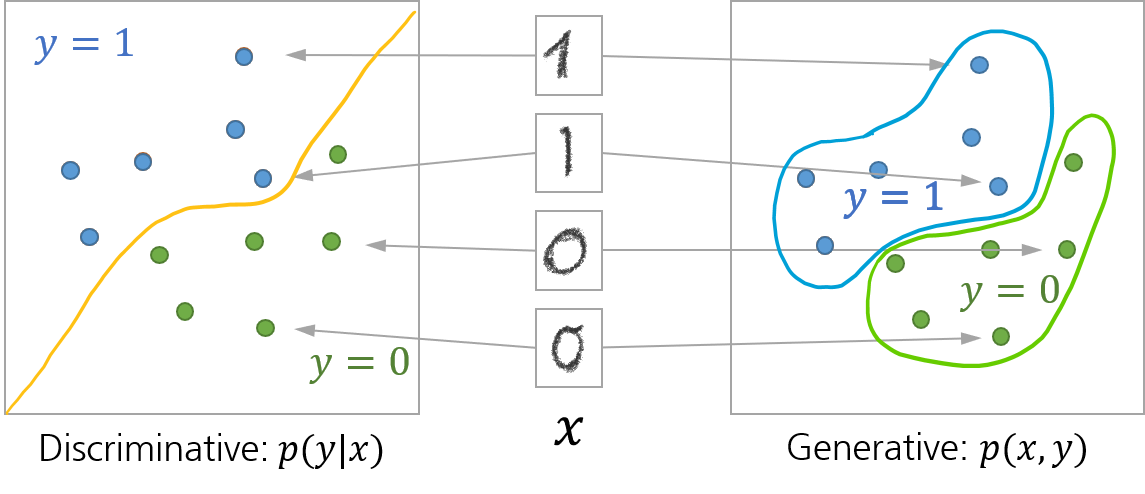}
	   \caption[Discriminative vs. generative task]{The discriminative task compared to the generative task. Discriminative models only need to find the separating line between classes, while generative models need to delineate the part of space covering the classes (Figure inspired by: https://developers.google.com/machine-learning/gan/generative).}
	\label{fig:disc-vs-gen}
\end{figure}

Generative models are fundamentally different, in that they generate output potentially without any concrete input, out of randomness. Still, they are supposed to generate output that conforms to certain criteria. In the most general form and intuitive formulation, their output should ``look natural''. We want to further formalize the difference between the models in the following by using the perspective of distributions again. \autoref{fig:disc-vs-gen} shows how discriminative and generative models have to construct differently complex boundaries in the representation space of the domain to accomplish their tasks.

Discriminative models take one example and map it to a label – e.g. the class. This is also true for segmentation models: they do this for each image voxel. The conceptual process is that the model has to estimate the probabilities that the example (or the voxel) comes from the distribution of the different available classes. The distributions of all possible appearances of objects of all classes do not need to be modelled analytically for this to be successful. It is only important to know them locally – for example it is sufficient to delineate their borders or overlaps with other distributions of other classes, but not all boundaries are important.

Generative models, on the other hand, are tasked to produce an example that is within a desired distribution. For this to work, the network has to learn the complete shape of this distribution. This is immensely complex, since all domains of practical importance in medical imaging are extremely high dimensional, and the distributions defining examples of interest within these domains are very small and hard to find. Also, they are neither analytically given nor normally distributed in their multi-dimensional space . But they have as many parameters as the output image of interest has voxels.

As already remarked, different other approaches were devised to generate output before GANs entered the scene. Among the trainable ones, approaches comprised (Restricted) Boltzman machines, Deep Belief Networks, or Generative Stochastic Networks, Variational Autoencoders, and others. Some of them involved feedback loops in the inference process (the prediction of a generated example), and were therefore unstable to train using backpropagation. 

This was solved with the adversarial net framework proposed in 2014 by \citeauthor{Goodfellow2014GAN}~\cite{Goodfellow2014GAN}. They tried to solve the downsides like computational intractability or instability of such previous generative models by introducing the adversarial training framework. 

To understand how GANs relate to one of the closest predecessors, the Variational Autoencoder, we will review their basic layout next. We will learn how elegantly the GAN paradigm turns the previously unsupervised approach to generative modeling into a supervised one, with the benefit of much more control over the training process. 

\subsection{Before GANs: Variational Autoencoders}
\label{sec:vae}

Generative Adversarial Networks (GANs) haven't been the first or only attempt at generating realistically-looking images (or any type of output, generally speaking). Apart from GANs, a related neural-network-based approach to generative modeling is the Variational Autoencoder, which will be treated in more detail below. Among other generative models with different approaches are:

\begin{description}
    \item[Flow-based models.]{This category of generative models attempt to model the data-generating distribution explicitly through an iterative process known as the Normalizing Flow~\cite{JimenezRezende2015VariationalIW}, in which through repeated changes of variables a sequence of differentiable basis distributions is stacked to model the target distribution. The process is fully invertible, yielding models with desirable properties, since an analytical solution to the data-generating distribution allows to directly estimate densities to predict the likelihood of future events, impute missing data points, and of course generate new samples. Flow-based models are computation-intensive. They can be categorized as a method that returns an explicit, tractable density. Another method in this category is for example the PixelRNN~\cite{Oord2016PixelRN} or the PixelCNN~\cite{Oord2016ConditionalIG} which also serves for conditional image generation. RealNVP~\cite{Dinh2017RealNVP} also uses a chain of invertible functions.}
    \item[Boltzmann machines]{ work fundamentally differently. 
    They also return explicit densities, but this time only approximate the true target distribution. In this regard, they are similar to Variational Autoencoders, though their method is based on Markov Chains, and not a variational approach. Deep Boltzmann Machines have been proposed already in 2009, uniting a Markov Chain-based loss component with a Maximum-Likelihood-based component and showing good results on, at that time, highly complex datasets.~\cite{pmlr-v5-salakhutdinov09a} Boltzmann machines are very attractive, but harder to train and use than other comparably powerful alternatives that exist today. This might change with future research, however.}
\end{description}

\begin{figure}[hbtp]
	\centering
		\includegraphics[width=0.9\textwidth]{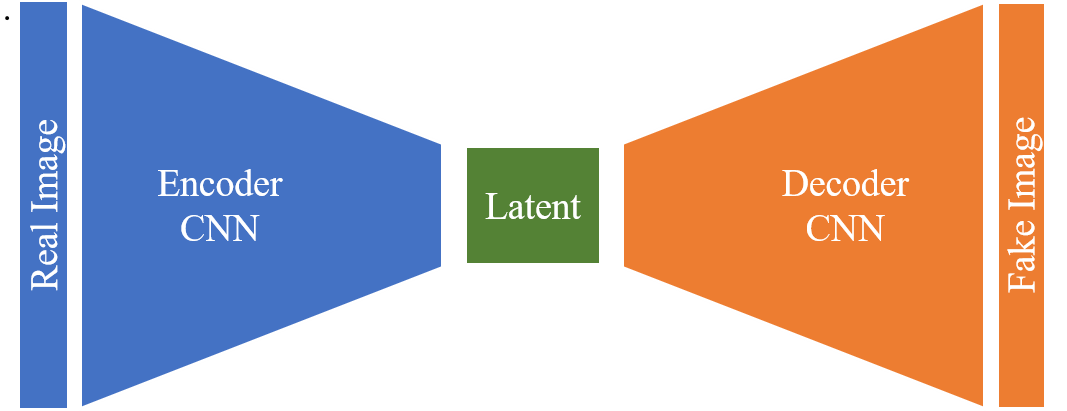}
	   \caption[Autoencoder Schematic]{Schematic of an Autoencoder network. The encoder, for images for example a CNN with a number of convolutional and pooling layers, condenses the defining information of the input image into the variables of the latent space. The decoder, again convolutions, but this time with up-sampling layers, recreates a representation in image space. Input and output images are compared in the loss function, which drives the gradient descent.}
	\label{fig:ae}
\end{figure}

Variational Autoencoders (VAE) are a follow-up development of plain Autoencoders, autoregressive models that in their essence try to reconstruct their input after transforming it, usually into a low-dimensional representation (see \autoref{fig:ae}). This low-dimensional representation is often termed the ``latent space'', implying that here hidden traits of the data-generating process are coded, that are essential to the reconstruction process. This is very akin to the latent variables estimated by EM. In the Autoencoder, the encoder will learn to code its input in terms of these latent variables, while the decoder will learn to represent them again in the source domain. In the following, we will be discussing the application to images though, in principle, both Autoencoders and their variational variant are general mechanisms working for any domain.

We will later be interested in a behind-the-scenes understanding of their modeling approach, which will be related to the employed loss function. We will then look at VAEs more extensively from the same vantage point: to understand their loss function -- which is closest to the loss formulation of early GANs: the Kullback-Leibler divergence, or KL divergence, $D_\mathrm{KL}$. 

With this tool in hand, we will examine how to optimize (train) a network with regard to KL divergence as the loss and understand key problems with this particular loss function. This will lead us to the motivation for a more powerful alternative.

\subsubsection{From AE to VAE}
VAEs are an interesting subject to study to emphasize the limits a loss function like KL divergence may place on a model. We will begin with a recourse to plain Autoencoders to introduce the concept of learning a latent representation. We will then proceed to modify the autoencoder into a variational formulation which brings about the switch to a divergence measure as a loss function. From these grounds, we will then show how GANs again modified the loss function to succeed in high-quality image generation.

\autoref{fig:ae} shows the schematic of a plain autoencoder (AE). As indicated in the sketch, input and output are of potentially very high dimensionality, like images. In between the encoder and decoder networks lies a ``bottleneck'' representation, which is for example a convolutional layer of orders of magnitude lower dimensionality (represented for example by a convolutional layer with only a few channels, or a dense layer with a given low number of weights), that forces the network to find an encoding that preserves all information required for reconstruction. 

A typical loss function to use when training the autoencoder is for example cross-entropy, which is applicable for sigmoid activation functions, or simply the mean squared error (MSE). Any loss shall essentially force the AE to learn the identity function between input and output. 

Let us introduce the notation for this. Let $\bm{\mathsfit{X}}$ be the input image tensor and $\bm{\mathsfit{X'}}$ the output image tensor. With $f_w$ being the encoder function given as a neural network parameterized by weights and biases $w$ and $g_v$ the decoder function parameterized by $v$, the loss hence works to make $\bm{\mathsfit{X}}=\bm{\mathsfit{X'}}=g_v(f_w(\bm{\mathsfit{X}}))$. 

In a \emph{variational} autoencoder\footnote{Though variational autoencoders are in general not necessarily neural networks, in our context we restrict ourselves to this implementation and stick to the notation with parameters $w$ and $v$, where in many publications they are denoted $\theta$ and $\phi$.}, things work differently. Autoencoders like before use a fixed (deterministic) latent code to map the input to, while Variational Autoencoders will replace this with a distribution. We can call this distribution $p_w$, indicating the parameterization by $w$. It is crucial to understand that a choice was made here that imposes conditions on the latent code. It is meant to represent the input data in a variational way: in a way following Bayes' laws. Our mapping of the input image tensor $\bm{\mathsfit{X}}$ to the latent variable $\mathbf{z}$ is by this choice defined by 
\begin{itemize}
    \item the prior probability $p_w(\mathbf{z})$, 
    \item the likelihood (conditional probability) $p_w(\bm{\mathsfit{x}} \vert \mathbf{z})$, 
    \item and the posterior probability $p_w(\mathbf{z}\vert \bm{\mathsfit{x}})$.
\end{itemize}

Therefore, once we have obtained the correct parameters $\hat{w}$ by training the VAE, we can produce a new output $\bm{\mathsfit{X}}'$ by sampling a $\mathbf{z}^{(i)}$ from the prior probability $p_{\hat{w}}(\mathbf{z})$ and then generate the example from the conditional probability through $\bm{\mathsfit{X}}^{(i)}=p_{\hat{w}}(\bm{\mathsfit{X}}|\mathbf{z}=\mathbf{z}^{(i)})$.

Obtaining the optimal parameters, however, isn't possible directly. The searched optimal parameters are those that maximize the probability that the generated example $\bm{\mathsfit{X}}'$ looks real. This probability can be rewritten as the aggregated conditional probabilities: 

$$p_w (\bm{\mathsfit{X}}^{(i)}) = \int p_w(\bm{\mathsfit{X}}^{(i)}\vert\mathbf{z}) p_w(\mathbf{z}) d\mathbf{z}.$$

\begin{figure}[hbtp]
	\centering
		\includegraphics[width=0.9\textwidth]{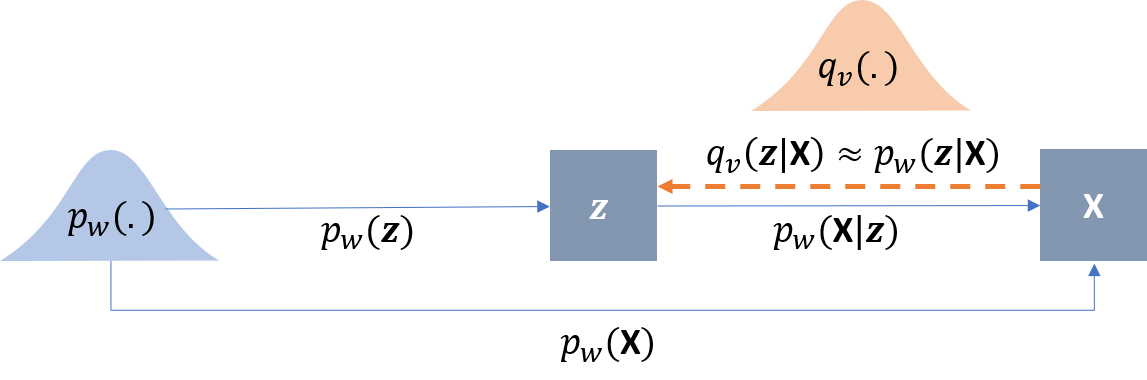}
	   \caption[Variational Autoencoder Graphical Model]{The graphical model of the Variational Autoencoder. In a VAE, the \emph{variational decoder} is $p_w(\bm{\mathsfit{X}}|\mathbf{z})$, while the \emph{variational encoder} is $q_v(\mathbf{z}|\bm{\mathsfit{X}})$. (Figure after \cite{weng2018VAE})}
	\label{fig:vae-graph-model}
\end{figure}

This, however, does not make the search any easier since we need to enumerate and sum up all $\mathbf{z}$. Therefore, an approximation is made through a surrogate distribution, parameterized by another set of parameters, $q_v$. \citeauthor{weng2018VAE} \cite{weng2018VAE} shows in her explanation of the VAE  the graphical model highlighting how $q_v$ is a stand-in for the unknown searched $p_w$ (see \autoref{fig:vae-graph-model}). 

The reason to introduce this surrogate distribution actually comes from our wish to train neural networks for the decoding/encoding functions, and this requires us to back-propagate through the random variable, $\mathbf{z}$, which of course cannot be done. Instead, if we have control over the distribution, we can select it such that the reparameterization trick can be employed. We define $q_v$ to be a multivariate Gaussian distribution with means and a covariance matrix that can be learned, and a stochastic element multiplied to the covariance matrix for sampling~\cite{kingma2014autoencoding,weng2018VAE}. With this, we can back-propagate through the sampling process.

At this point, the two distributions need to be made to match: $q_v$ should be as similar to $p_w$ as possible. Measuring their similarity can be done in a variety of ways, of which Kulback-Leibler divergence (KL divergence or KLD) is one.

\subsubsection{KL Divergence}

A divergence can be thought of as an asymmetric distance function between two probability distributions, $P$ and $Q$, measuring the similarity between them. It is a statistical distance which is not symmetric, which means it will not yield the same value if measured from $P$ to $Q$ or the other way around:

$$D_\mathrm{KL}(P \Vert Q) \ne D_\mathrm{KL}(Q \Vert P)$$

This can be seen when looking at the definition of KL divergence:

\begin{equation}
  D_\mathrm{KL}(P \Vert Q) = \sum_x {P(x) \log \frac{P(x)}{Q(x)}} 
  \label{eqn:kld}
\end{equation}

Sometimes, the measure $D_\mathrm{KL}$ is also called the relative entropy or information gain of $P$ over $Q$, which also indicates the asymmetry. 

To give the two distributions more meaning, let us associate them with a use case. $P$ is usually the probability distribution of the example data, which can be our real images we wish to model, and is assumed to be unknown and high-dimensional. $Q$, on the other hand, is the modeled distribution, for example parameterized by $\theta$, similar to~\autoref{eqn:max-likelihood}. Hence, $Q$ is the distribution we can play with (in our case: optimize its parameters) to make them more similar to $P$. This means, $Q$ will get more informative with respect to the true $P$ when we approach the optimal parameters. 

\begin{figure}[hbtp]
	\centering
		\includegraphics[width=0.95\textwidth]{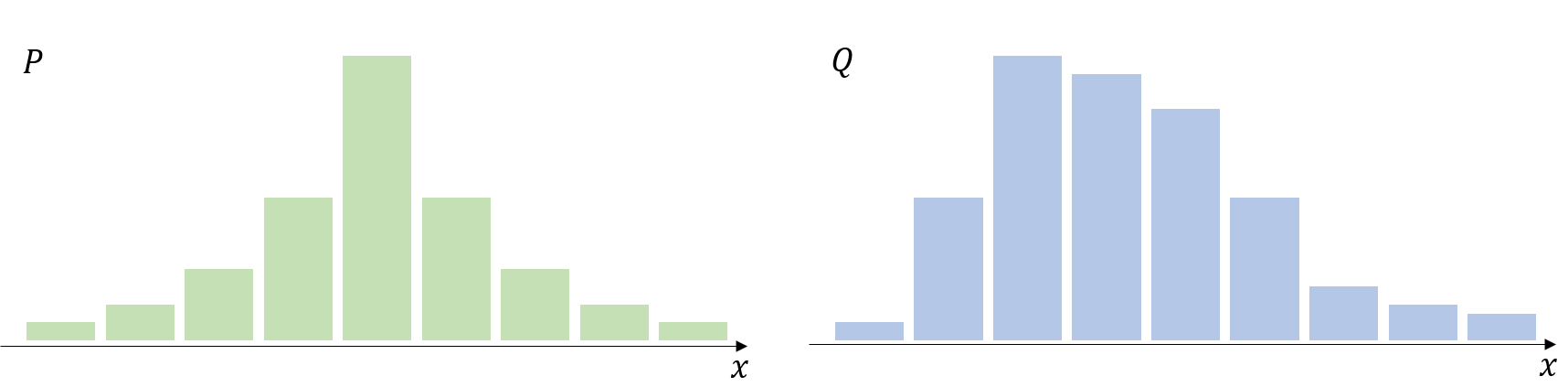}
	   \caption[Example distributions]{Two distributions $P$ and $Q$, here scaled to identical height.}
	\label{fig:kl-1}
\end{figure}

\begin{figure}[hbtp]
	\centering
		\includegraphics[width=0.65\textwidth]{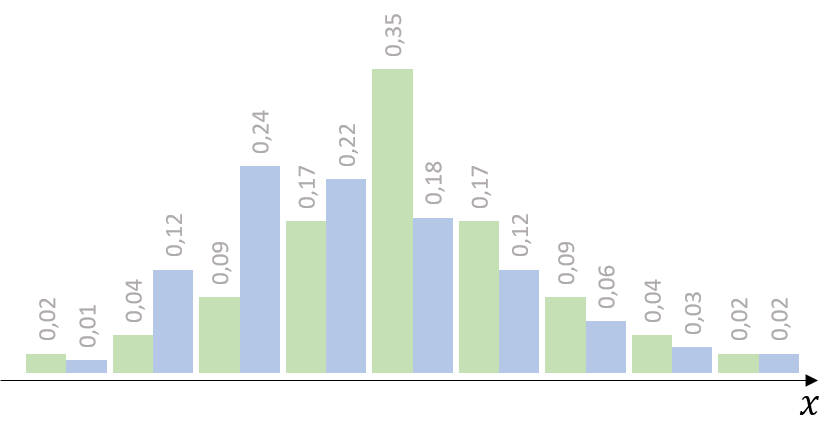}
	   \caption[KLD example]{The distributions $P$ and $Q$, scaled to unit density, with added labels.}
	\label{fig:kl-2}
\end{figure}

\begin{floatbox}[hbtp]
    \begin{nicebox}[Example: Calculating $D_\mathrm{KL}$]
        \label{box:KLD}
    
        When comparing the two distributions given in~\autoref{fig:kl-1}, the calculation of the Kullback-Leibler divergence, $D_\mathrm{KL}$, can explicitly be given by reading off the $y$ values of the nine elements (columns) from~\autoref{fig:kl-2} and inserting them into~\autoref{eqn:kld}.
        
        The result of this calculation is for 
        
        \begin{equation*}
        \begin{split}
          D_\mathrm{KL}(P \Vert Q) & = \sum_x {P(x) \log \frac{P(x)}{Q(x)}} \\
            & = .02*\log\frac{.02}{.01} + .04*\log \frac{.04}{.12} + \dots + .02*\log \frac{.02}{.022}\\ 
            & = .004 - .01 + \dots - .0002 \\
            & = .0801         
        \end{split}
        \end{equation*}
    
        which we call ``forward KL'' as it calculates in the direction from the actual distribution $P$ to the model distribution $Q$; and for 
        
        \begin{equation*}
        \begin{split}
          D_\mathrm{KL}(Q \Vert P) & = \sum_x {Q(x) \log \frac{Q(x)}{P(x)}} \\
            & = .01*\log\frac{.01}{.02} + .12*\log \frac{.12}{.04} + \dots + .022*\log \frac{.022}{.02}\\ 
            & = -.002 - .05 + \dots + .0002 \\
            & = .0899         
        \end{split}
        \end{equation*}
    
        which we call ``reverse KL''.
    \end{nicebox}
\end{floatbox}

Note that in the example in~\autoref{box:KLD}, there is both a $P(X=x_i)$ and $Q(X=x_i)$ for each $i\in \{0, 1, \ldots, 8 \}$. This is crucial for KL divergence to work as a loss function. 

\subsubsection{Optimizing the KL Divergence}

Examine what happens in forward and reverse KL if this condition is not satisfied for some $i$. If in forward KL, $P$ has values everywhere, but $Q$ has not (or extremely small values), the quotient in the $\log$ function will tend to infinity by means of the division by almost zero, and the term will be very large. 

\begin{figure}[hbtp]
	\centering
		\includegraphics[width=0.95\textwidth]{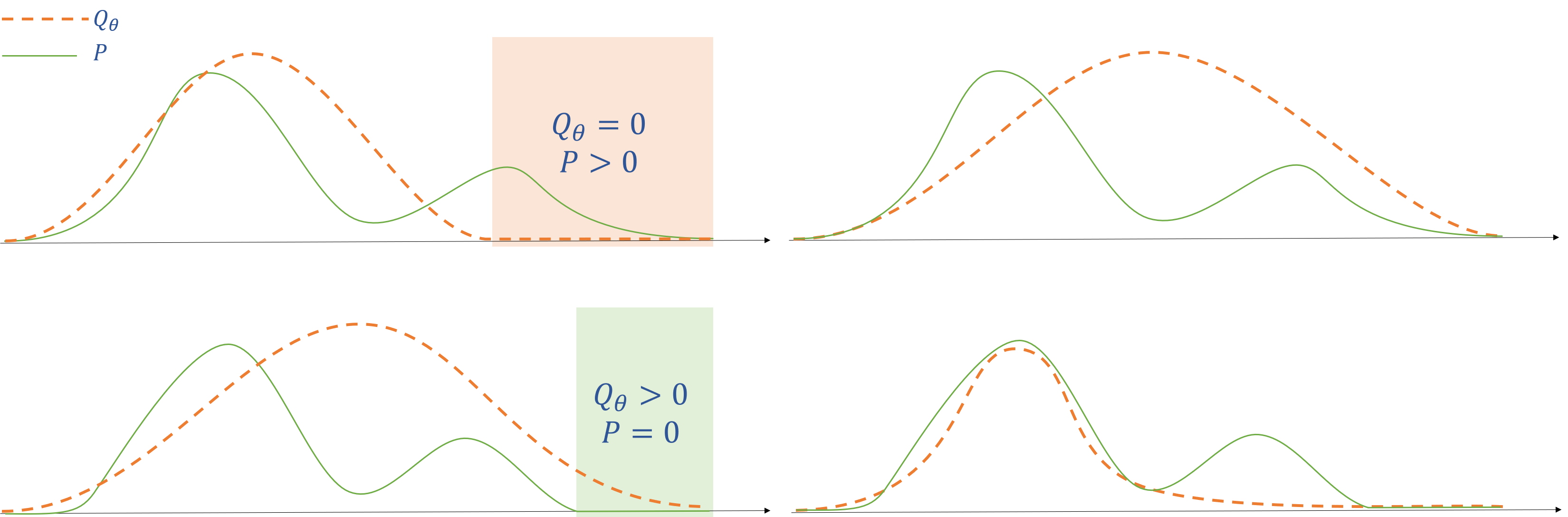}
	   \caption[Forward and Reverse KL]{The distributions $P$ (solid) and $Q_\theta$ (dashed), in the initial configuration and after minimizing reverse KL $D_\mathrm{KL}(Q_\theta \vert P)$. This time, in the initial configuration, $Q_\theta$ has values greater than 0 where $P$ has not (marked with green shading).}
	\label{fig:forward-reverse-kl}
\end{figure}

In~\autoref{fig:forward-reverse-kl}, we assume $Q_\theta$ to be a uni-modal normal distribution, i.e. a Gaussian, while $P$ is any empirical distribution. In the left plots of the Figure, we show a situation before minimizing the forward/reverse KL divergence between $P$ and $Q_\theta$, in the right plots, the resulting shape of the Gaussian after minimization. 



When in the minimization of forward KL $D_\mathrm{KL}(P \vert Q_\theta)$, $Q_\theta$ is zero where $P$ has values greater zero, KL goes to infinity in these regions (marked area in the start configuration of the top row in~\autoref{fig:forward-reverse-kl}), since the denominator in the $\log$ function goes to zero. This, in turn, drives the parameters of $Q_\theta$ to broaden the Gaussian to cover these areas, thereby removing the large loss contributions. This is known as the \emph{mean-seeking} behaviour of forward KL.

Conversely, in reverse KL (bottom row in~\autoref{fig:forward-reverse-kl}), in the marked areas of the initial configuration, $P$ is zero in regions where $Q_\theta$ has values greater than zero. This yields high loss contributions from the $\log$ denominator, in this case driving the Gaussian to remove these areas from $Q_\theta$. Since we assumed a uni-modal Gaussian $Q$, the minimization will focus on the largest mode of the unknown $P$. This is known as the \emph{mode-seeking} behaviour of reverse KL.

Forward KL tends to overestimate the target distribution, which is exaggerated in the right plot in~\autoref{fig:forward-reverse-kl}. In contrast, reverse KL tends to underestimate the target distribution, for example by dropping some of its modes. Since underestimation is the more desirable property in practical settings, reverse KL is the loss function of choice for example in Variational Autoencoders. The downside is that, as soon as target distribution $P$ and model distribution $Q_\theta$ have no overlap, KL divergence evaluates to infinity and is therefore uninformative. One countermeasure to take is to add noise to $Q_\theta$, so that there is guaranteed overlap. This noise, however, is not desirable in the model distribution $Q_\theta$ since it disturbs the generated output.

Another way to remedy the problem of KL going to infinity is to adjust the calculation of the divergence, which is done in Jensen-Shannon divergence (JS divergence, $D_\mathrm{JS}$) defined as

\begin{equation}
    \label{eqn:jsd}
    D_\mathrm{JS} = \frac{1}{2} (D_\mathrm{KL}(P \Vert M) + D_\mathrm{KL}(Q_\theta \Vert M)),
\end{equation}

where $M = \frac{P+Q_\theta}{2}$. In the case of non-overlapping $P$ and $Q_\theta$, this evaluates to constant $\log 2$, which is still not providing information about the closeness, but is computationally much friendlier and does not require the addition of a noise term to achieve numerical stability.

\subsubsection{The Limits of VAE}
In the VAE, reverse KL is used. Our optimization goal is maximizing the likelihood to produce realistic looking examples -- ones with a high $p_w(x)$. Simultaneously, we want to minimize the difference between the real and estimated posterior distributions $q_v$ and $p_w$. This can only be achieved through a reformulation of reverse KL~\cite{weng2018VAE}. After some rearranging of reverse KL, the loss of the Variational Autoencoder becomes

\begin{equation}
    \begin{aligned}
    L_\mathrm{VAE}(w,v) 
     & = -\log p_w(\bm{\mathsfit{X}}) + D_\mathrm{KL}( q_v(\mathbf{z}\vert\bm{\mathsfit{X}}) \| p_w(\mathbf{z}\vert\bm{\mathsfit{X}}) )\\
     & = - \mathbb{E}_{\mathbf{z} \sim q_v(\mathbf{z}\vert\bm{\mathsfit{X}})} \log p_w(\bm{\mathsfit{X}}\vert\mathbf{z}) + D_\mathrm{KL}( q_v(\mathbf{z}\vert\bm{\mathsfit{X}}) \| p_w(\mathbf{z}) )
    \end{aligned}
\end{equation}

$\hat{w}, \hat{v}$ are the parameters maximizing the loss.

We have seen how mode-seeking reverse KL divergence limits the generative capacity of Variational Autoencoders through the potential under-representation of all modes of the original distribution. 

KL divergence and minimizing the ELBO also has a second fundamental downside: there is no way to find out how close our solution is to the obtainable optimum. We measure the similarity to the target distribution up to the KL divergece, but since the true $p_{\hat{w}}(.)$ is unknown, the stopping criterion in the optimization has to be set by another metric, e.g. to a maximum number of iterations or corresponding to an improvement of the loss below some $\varepsilon$.

The original presentation of the Variational Autoencoder was given as one example of the general framework called the Autoencoding Variational Bayes. This publication presented the above ideas in a thorough mathematical formulation, starting from a directed graphical model that poses the abstract problem. The authors also develop the seminal ``reparameterization trick'' to make the loss formulation differentiable, and with this to make the search for the autoencoder parameters amenable to gradient descent optimizers~\cite{kingma2014autoencoding}. The details are beyond this introductory treatment.

\subsection{The fundamental GAN approach}
\label{ssec:basic_gan}

At the core of the adversarial training paradigm is the idea to create two players competing in a minimax game. In such games, both players have access to the same variables but have opposing goals, so that they will manipulate the variables in different directions. 

\begin{figure}[hbtp]
	\centering
		\includegraphics[width=0.95\textwidth]{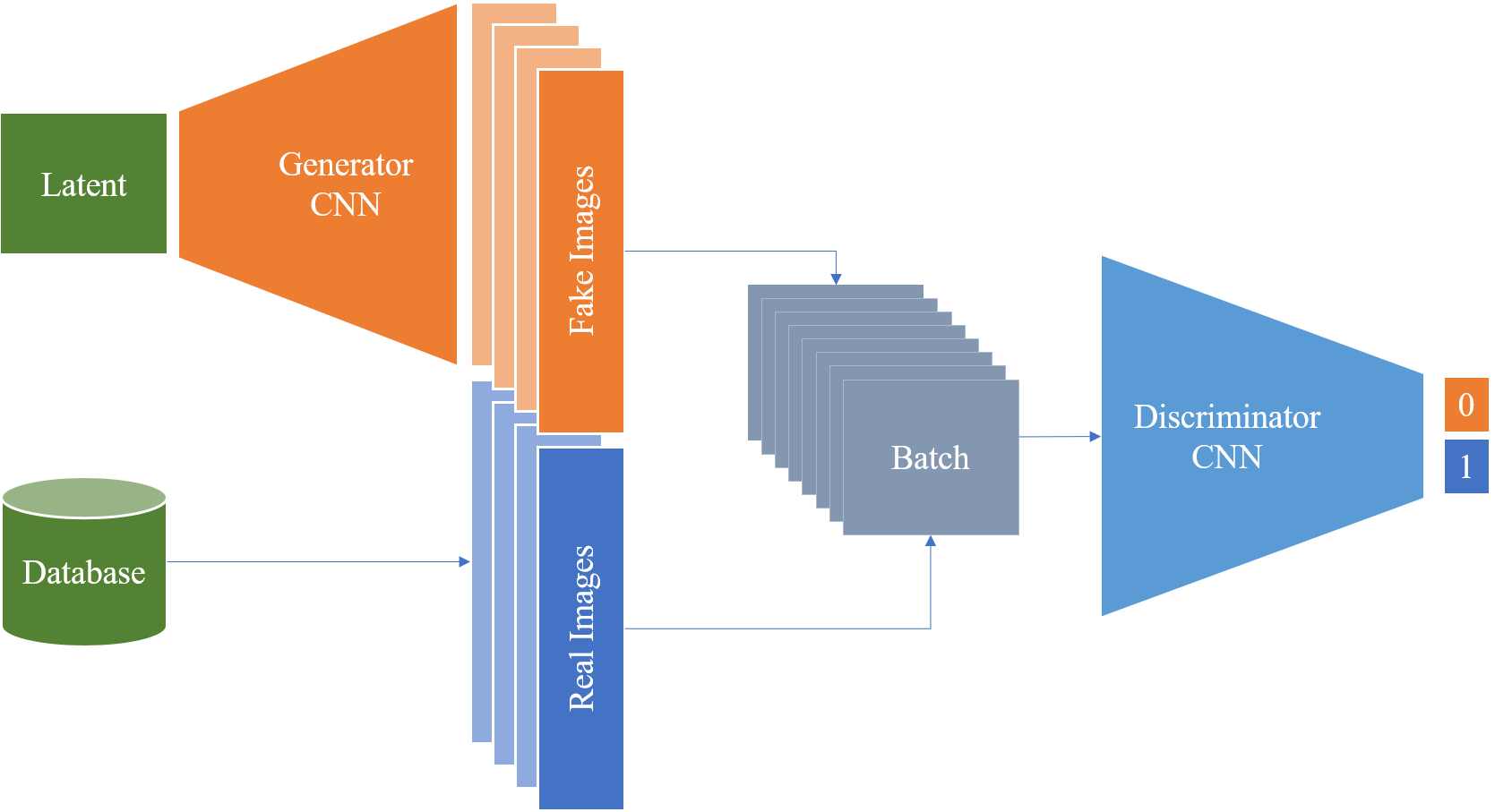}
	   \caption[GAN Schematic]{Schematic of a GAN network. Generator (orange) creates fake images based on random numbers drawn from a latent space. These together with a random sample of real images are fed into the discriminator (blue, right). The discriminator looks at the batch of real/fake images and tries to assign the correct label (``0'' for fake, ``1'' for real).}
	\label{fig:basicgan}
\end{figure}

Referring to \autoref{fig:basicgan}, we can see the generative part in orange color, where random numbers are drawn from the latent space and, one by one, converted into a set of ``fake images'' by the generator network, in the Figure implemented by a CNN. Simultaneously, from a database of real images a matching number of examples is randomly drawn. The real and fake images are composed into one batch of images which are fed into the discriminator. On the right side, the Discriminator CNN is indicated in blue. It takes the batch of real and fake images and decides for each if it appears real (yielding a value close to ``1'') or fake (``0''). 

The error signal is computed from the number of correct assignments the discriminator can do on the batch of generated and real images. Both the generator and the discriminator can then update their parameters based on this same error signal. Crucially, the generator has the aim to \emph{maximize} the error, since this signifies that it has successfully fooled the discriminator into taking the fake images for real, while the discriminator weights are updated to \emph{minimize} the same error, indicating its success in telling true and fake examples apart. This is the core of the competitive game between generator and discriminator.

Let us introduce some abbreviations to designate GAN components. We will denote the generator and discriminator networks with $G$ and $D$, respectively. The objective of GAN training is a game between generator and discriminator, where both affect a common loss function $J$, but in opposed directions. Formally, this can be written as

\begin{equation*}
    \min_G\max_D J(G, D),
\end{equation*}

with the GAN objective function 

\begin{equation}
    \label{eqn:gan-objective}
    J(G, D) = \mathbb{E}_{x \sim p_\mathrm{data}}[\log D(x)] + \mathbb{E}_{z \sim p_G}[1 - \log D(G(z))]
\end{equation}

$D$ will attempt to maximize $J$ by maximizing the probability to assign the correct labels to real and generated examples: this is the case if $D(x)=1$, maximizing the first loss component, and if $D(G(z))=0$, maximizing the second loss component. The generator $G$, instead, will attempt to generate realistic examples that the discriminator labels with ``1'', which corresponds to a minimization of $\log(1-D(G(z)))$.

\subsection{Why early GANs were hard to train}
\label{ssec:early_problems}

GANs with this training objective implicitly use JS divergence for the loss, which can be seen by examining the GAN training objective. Consider the ideal discriminator $D$ for a fixed generator. Its loss is minimal for the optimal discriminator given by~\cite{Goodfellow2014GAN}:

\begin{equation}
  \hat{D}(x) = \frac{p_\mathrm{data}(x)}{p_\mathrm{data}(x) + p_G(x)}. 
\end{equation}

Substituting $\hat{D}$ in~\autoref{eqn:gan-objective} yields (without proof) the implicit use of the Jensen-Shannon divergence if the above training objective is employed:

\begin{equation}
    \label{eqn:gan-js}
    J(G, \hat{D}) = 2D_\mathrm{JS}(p_\mathrm{data} \Vert p_G) - \log 4.
\end{equation}

This theoretical result shows that a minimum in the GAN training can be found when the Jensen-Shannon divergence is zero. This is achieved for identical probability distributions $p_\mathrm{data}$ and $p_G$, or equivalently, when the generator perfectly matches the data distribution~\cite{Creswell2018GANOverview}. 

Unfortunately, it also shows that this loss is, like KL divergence, only helpful when target distribution (i.e. data distribution) and model distribution have overlapping support. Therefore, added noise can be required to approximate the target distribution. In addition, the training criterion saturates if the discriminator in the early phase of training perfectly distinguishes between fake and real examples. The generator will therefore no longer obtain a helpful gradient to update its weights. An approach thought to prevent this was proposed by \citeauthor{Goodfellow2014GAN}~\cite{Goodfellow2014GAN}. The generator loss was turned from the minimization problem into a maximization problem that has the same fixed point in the overall minimax game, but prevents saturation: instead of minimizing $\log(1-D(G(z)))$, one maximizes $\log(D(G(z)))$~\cite{Goodfellow2014GAN}.

\subsection{Improving GANs}
\label{ssec:improving}

GAN training has quickly become notorious for the difficulties it posed upon the researchers attempting to apply the mechanism to real-world problems. We have qualitatively attributed a part of these problems to the inherently difficult task of density estimation, and motivated the intuition that while fewer samples might suffice to learn a decision boundary in a discriminative task, many more examples are required to build a powerful generative model.

In the following, some more light shall be shed on the reasons why GAN training might fail. Typical GAN problems comprise:

\begin{description}
    \item [Mode dropping] is the phenomenon in forward KL caused by regions of the data distribution not being covered by the generator distribution, which implies large probabilities of samples coming from $P_\mathrm{data}$, and very small probabilities of originating from $P_G$. This drives forward KL towards infinity and punishes the generator for not covering the entire data distribution~\cite{Arjovsky2017TowardsPM}. If all modes but one are dropped, one can call this mode collapse: the generator only generates examples from one mode of the distribution.
    \item [Poor convergence] can be caused by a discriminator learning to distinguish real and fake examples very early -- which is also very likely to happen throughout the GAN training. This is rooted in the observation that by the generative process that projects from a low-dimensional latent space into the high-dimensional $p_G$, the samples in $p_G$ are not close to each other, but rather inhabit ``islands''~\cite{Arjovsky2017TowardsPM}. The discriminator can learn to find them, and thereby differentiate between true and false samples easily, which causes the gradients driving generator optimization to vanish~\cite{Creswell2018GANOverview}. 
    \item [Poor sample quality] despite a high log likelihood of the model is a consequence of the practical independence of sample quality and model log likelihood. \citeauthor{Theis2016ANO} \cite{Theis2016ANO} show that neither does a high log likelihood imply generated sample fidelity, nor do visually pleasing samples imply a high log likelihood. Therefore, training a GAN with a loss function that effectively implements maximizing a log likelihood term is not an ideal choice -- but exactly corresponds to KL minimization. 
    \item [Unstable Training] is a consequence of reformulating the generator loss into maximizing $\log D(G(z))$. It can be shown~\cite{Arjovsky2017TowardsPM} that this choice effectively makes the generator struggle between a reverse KL divergence favouring mode-seeking behaviour, and a negative JS divergence actually driving the generator into examples different from the real data distribution.
\end{description}

There have been many subsequent authors touching these topics, but already \citeauthor{Arjovsky2017TowardsPM} \cite{Arjovsky2017TowardsPM} have shown best practices of how to overcome these problems. 

Among the solutions proposed for GAN improvements are some that prevent the generator from producing only too similar samples in one batch, some that keep the discriminator insecure about the true labels of real and fake examples, and more, which \citeauthor{Creswell2018GANOverview} \cite{Creswell2018GANOverview} have summarized in their GAN overview. A collection of best practices compiled from these sources is presented in \autoref{box:stableGAN}. It is almost impossible to write a cookbook for successful, converging, stable GAN training. For almost every tip, there is a caveat or situation where it cannot be applied. The suggestions below therefore are to be taken with a grain of salt, but have been used by many authors successfully.

\begin{floatbox}[hbtp]
    \begin{nicebox}[Best practices for stable GAN training]
        \label{box:stableGAN}

        \footnotesize 
        
        \textbf{General measures.}
        GAN training is sensitive to hyperparameters, most importantly the learning rate. Mode collapse might already be mitigated by a lower learning rate. Also, different learning rates for generator and discriminator might help. Other typical measures are batch normalization (or instance normalization in case of small batch sizes; mind however that batch normalization can taint the randomness of latent vector sampling, and in general should not be used in combination with certain GAN loss functions), use of transposed convolutions instead of parameter-free up-sampling, strided convolutions instead of down-sampling.
        \vspace{6pt}
        
        \textbf{Feature matching.} 
        One typical observation is that neither discriminator nor generator converge. They play their ``cat and mouse'' game too effectively. The generator produces a good image, but the discriminator learns to figure it out, and the generator shifts to another good image, and so on. 
        
        A remedy for this is feature matching, where the $\ell_2$ distance between the average feature vectors of real and fake examples is computed instead of a cross-entropy-loss on the logits. Because per batch the feature vectors change slightly, this introduces randomness that helps to prevent discriminator overconfidence.
        \vspace{6pt}

        \textbf{Minibatch discrimination.}
        When the generator only produces very convincing, but extremely similar images, this is an indication for mode collapse.
        
        This can be counteracted by calculating a similarity metric between generated samples, and penalizing the generator for too little variation. Minibatch discrimination is considered to be superior in performance to Feature matching.
        \vspace{6pt}
        
        \textbf{One-sided label smoothing.}
        Deep classification models often suffer from overconfidence, focusing on only very few features to classify an image. If this happens in a GAN, the generator might figure this out and only produce the feature the discriminator uses to decide for a real example.
        
        A simple measure to counteract this is to not provide a ``1'' as a label for the real images in the batch, but a lower value. This way, the discriminator is penalized for overconfidence (when it returns a value close to ``1'').
        \vspace{6pt}
        
        \textbf{Cost function selection.}
        Several sources list possible GAN cost functions. Randomly trying them one by one might work but often some of the above measures, in particular learning rate and hyperparameter tuning, might be more successful first steps.
    \end{nicebox}
\end{floatbox}

Besides these methods, one area of discussion concerned the question if there is a need of balancing discriminator and generator learning and convergence at all. The argument was that a converged discriminator will as well yield a training signal to the generator as a non-converged discriminator. Practically, however, many authors described carefully designed update schedules, e.g. updating the generator once per a given number of discriminator updates.

Many more ideas exist: weight updating in the generator using an exponential moving average of previous weights to avoid ``forgetting''; different regularization and conditioning techniques; injecting randomness into generator layers anew. Some we will encounter later, as they have proven to be useful in more recent GAN architectures.

\begin{figure}[hbtp]
	\centering
		\includegraphics[width=0.95\textwidth]{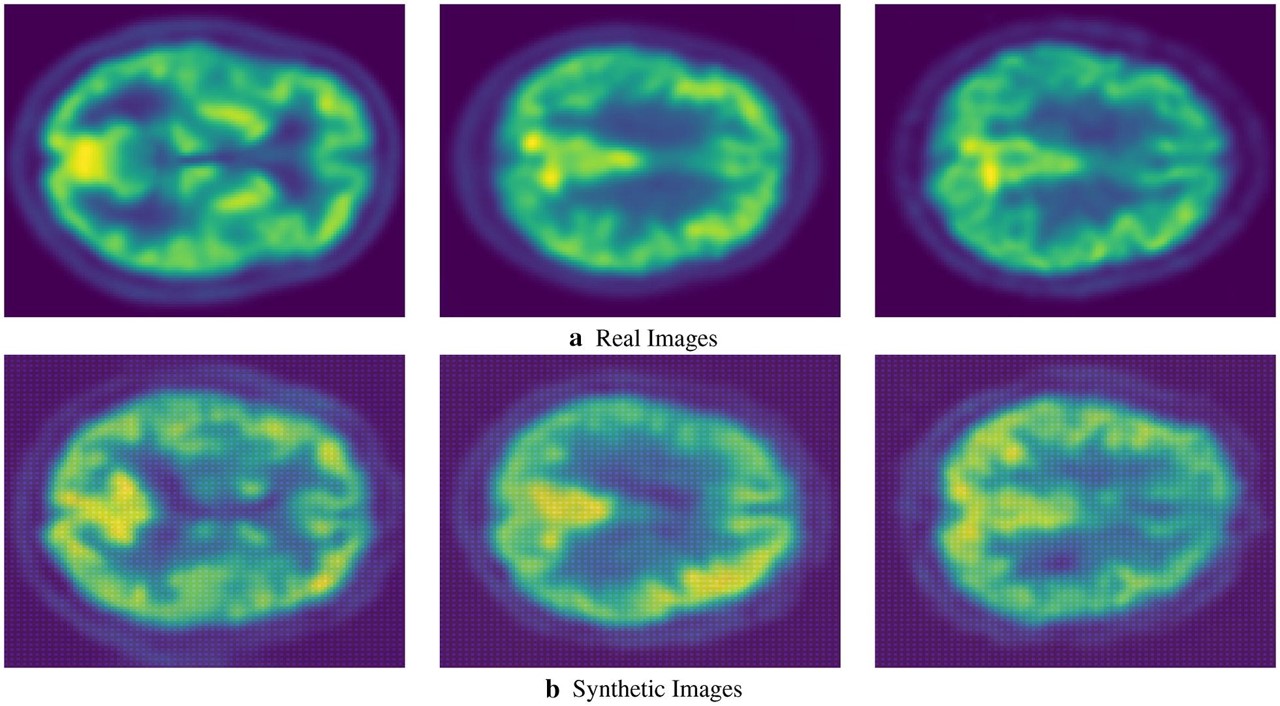}
	   \caption[Generated PET images]{PET images generated from random noise using a DCGAN architecture. Image taken from~\cite{Islam2020} (CC-BY4.0).}
	\label{fig:DCGANPET}
\end{figure}

Despite the recent advances in stabilizing GAN training, even the basic method described so far, with the improvements made in the seminal DCGAN publication~\cite{Radford2015}, finds application until today, e.g. for the de-novo generation of PET color images~\cite{Islam2020}. The usefulness of an approach as presented in their publication might be doubted, since the native PET data is obviously not colored. The authors use 2D histograms of the three color channel combinations to compare true and fake examples. As we have discussed earlier, this is likely a poor metric since it does not allow insights into the high-dimension joint probability distribution underlying the data generating process. \autoref{fig:DCGANPET} shows an example comparison of some generated examples compared to original PET images.

To address many of the GAN training dilemmas, \citeauthor{Arjovsky2017TowardsPM} have proposed to employ the Wasserstein distance as a replacement for KL or JS divergence already in their examination of the root causes of poor GAN training results, and have later extended this into their widely anticipated approach we will focus on next \cite{Arjovsky2017c,Gulrajani2017a}. We will also see more involved and recent approaches to stabilize and speed up GAN training in later sections of this chapter (\autoref{sec:architectures}).

\subsection{Wasserstein GANs}
\label{sec:wgan}

Wasserstein GANs were appealing to the deep learning and GAN scene very quickly after \citeauthor{Arjovsky2017c}'s~\cite{Arjovsky2017c} seminal publication because of a number of traits their inventors claimed they'd have. For one, Wasserstein GANs are based on the theoretical idea that the change of the loss function to the Wasserstein distance should lead to improved results. This combined with the reported benchmark performance would already justify attention. But Wasserstein GANs additionally were reported to train much more stably, because, as opposed to previous GANs, the discriminator would be trained to convergence in every iteration, instead of demanding a carefully and heuristically found update schedule for generator and discriminator. In addition, the loss was directly reported to correlate with visual quality of generated results, instead of being essentially meaningless in a minimax game.

Wasserstein GANs are therefore worth an in-depth treatment in the following sections.

\subsubsection{The Wasserstein (Earth Mover) Distance}

\begin{figure}[hbtp]
	\centering
		\includegraphics[width=0.55\textwidth]{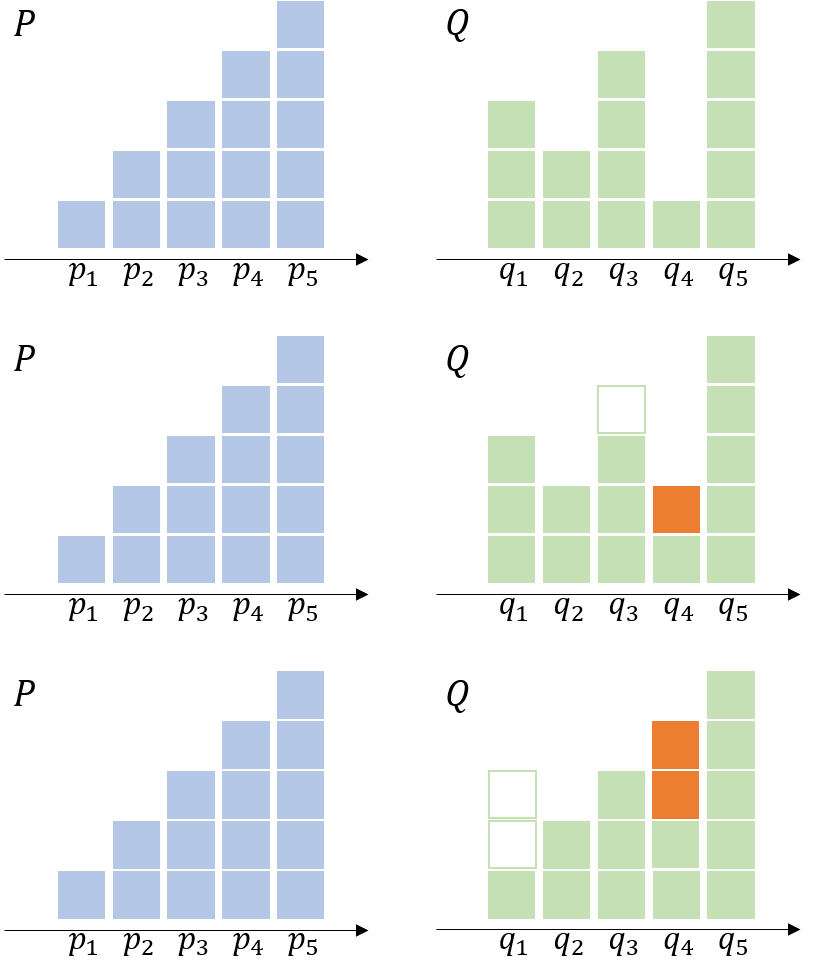}
	   \caption[Wasserstein: Moving Earth]{One square is one shovel full of earth. Transporting the earth shovel-wise from pile to pile amasses performed work: the Wasserstein (Earth Mover) distance. The example shows a Wasserstein distance of $D_\mathrm{W}=7$.}
    \label{fig:wassersteinbyhand}
\end{figure}

The Wasserstein distance figuratively measures how, with an optimal transport plan, mass can be moved from one configuration to another configuration with minimal work. Think for example of heaps of earth. \autoref{fig:wassersteinbyhand} shows two heaps of earth, $P$ and $Q$ (discrete probability distributions), both containing the same amount of earth in total, but in different concrete states $x$ and $y$ out of all possible states. 

Work is defined as the shovelfuls of earth times the distance it is moved. In the three rows of the Figure, earth is moved (only within one of $P$ or $Q$, not from one to the other), in order to make the configuration identical. First, one shovelful of earth is moved one pile further, which adds One to the Wasserstein distance. Then, two shovelfuls are moved three piles, adding six to the final Wasserstein distance of $D_\mathrm{W}=7$. 

Note that, in an alternative plan, it would have been possible to move two shovelfuls of earth from $p_4$ to $p_1$ (costing six), and one from $p_4$ to $p_3$, which is the inverse transport plan of the above, executed on $P$, and leading to the same Wasserstein distance. The Wasserstein distance is in fact a distance, not a divergence, because it yields the same result regardless of the direction. Also note that we implicitly assumed that $P$ and $Q$ share their support\footnote{The support, graphically, is the region where the distribution is not equal to zero.}, but that in case of disjunct support, only a constant term would have to be added, that grows with the distance between the support regions.

Many other transport plans are possible, and others can be equally cheap (or even cheaper -- it is left to the reader to try this out). Transport plans need not modify only one of the stocks, but can modify both to reach the optimal strategy to make them identical. Algorithmically, the optimal solution to the question of the optimal transport plan can be found by formulating it as a Linear Programming problem. However, enumerating all transport plans and computing the Linear Programming algorithm is intractable for larger and more complex ``heaps of earth''. Any non-trivial GAN will need to estimate transport of such complex ``heaps'', so they suffer this intractability problem. Consequently, in practice a different approach must be taken, which we will sketch below.\footnote{An extensive treatment of Wasserstein distance and optimal transport in general is given in the 1.000-pages treatment of \citeauthor{villani2009Transport}'s book~\cite{villani2009Transport}, which is freely available for download.}

Formalizing the search for the optimal transport plan, we look at all possible joint distributions of our $P$ and $Q$, forming the set of all possible transport plans, and denote this set $\Pi(P,Q)$, implying that for all $\gamma \in \Pi(P,Q)$, $P$ and $Q$ will be their marginal distributions.\footnote{This section owes to the excellent blog post of Vincent Herrmann, at \url{https://vincentherrmann.github.io/blog/wasserstein/}. Also recommended is the treatment of the ``Wasserstein GAN'' paper by Alex Irpan at \url{https://www.alexirpan.com/2017/02/22/wasserstein-gan.html}. An introductory treatment of Wasserstein distance is also found in~\cite{Basso2015,weng2019gan}.} 
This, in turn, means that by definition $\sum_x \gamma(x,y) = P(y)$ and $\sum_y \gamma(x,y) = Q(x)$. 

For one concrete transport plan $\gamma$ that works between a state $x$ in $P$ and a state $y$ in $Q$, we are interested in the optimal transport plan $\gamma(x,y)$. Let $\Vert x-y \Vert$ be the Euclidian distance to shift earth between $x$ and $y$, then multiplying this with every value of $\gamma$ (the amount of earth shifted) leads to

\begin{equation*}
    D_\mathrm{W}(P, Q) = \inf_{\gamma \in \Pi} \, \sum\limits_{x,y} \Vert x - y \Vert \gamma (x,y),
\end{equation*}

which can be rewritten to obtain
\begin{equation}
    \label{eqn:wasserstein_optimal_transport_plan}
    D_\mathrm{W}(P, Q) = \inf_{\gamma \sim \Pi(P, Q)} \mathbb{E}_{(x, y) \sim \gamma} \Vert x-y \Vert.
\end{equation}

It measures both the distance of two distributions with disjunct support and the difference between distributions with perfectly overlapping support because it includes both, the shifting of earth and the distance to move it.

Practically, though, this result cannot be used directly, since the Linear Programming problem scales exponentially with the number of dimensions of the domain of $P$ and $Q$, which are high for images. To our disadvantage, we additionally need to differentiate the distance function if we want to use it for deep neural network training using backpropagation. However, we cannot obtain a derivative from our distance function in the given form, since, in the Linear Programming (LP) formulation, our optimized distribution (as well as the target distribution) end up as constraints, not parameters. 

Fortunately, we are not interested in the transport plan $\gamma$ itself, but only in the distance (of the optimal transport plan). We can therefore use the dual form of the LP problem, in which the constraints of the primal form become parameters. With some clever definitions, the problem can be cast into the dual form, finally yielding

\begin{equation*}
    D_\mathrm{W}(P, Q) = \sup_{\lVert f \lVert_{L \leq 1}} \ \mathbb{E}_{x \sim P} f(x) - \mathbb{E}_{x \sim Q} f(x)
\end{equation*}

with a function $f$ that has to adhere to a constraint called the 1-Lipschitz continuity constraint, which requires $f$ to have a slope of at most magnitude 1 everywhere. $f$ is the neural network, and more specifically for a GAN, the discriminator network. 1-Lipschitzness can be achieved trivially by clipping the weights to a very small interval around 0.

\subsubsection{Implementing WGANs}
\label{sec:wgan-code}

To implement the distance as a loss function, we re-write the last result again as

\begin{equation}
    \label{eqn:wasserstein_loss}
    D_\mathrm{W}(P, Q) = \max_{w \in W} \mathbb{E}_{x \sim P}[D_w(x)] - \mathbb{E}_{z \sim Q}[D_w(G_w(z))].
\end{equation}

Note that in opposition to other GAN losses we have seen before, there is no logarithm anymore, because this time, the ``discriminator'' is no longer a classification network that should learn to discriminate true and fake samples, but rather serves as a ``blank'' helper function that during training learns to estimate the Wasserstein distance between the sets of true and fake samples.

\begin{floatbox}[htbp]
    \begin{nicebox}[Spectral normalization]
        \label{box:spectralnorm}
        
        Spectral normalization is applied to the weight matrices of a neural network to ensure a boundedness of the error function (e.g. Lipschitzness of the discriminator network in the WGAN context). This helps convergence like any other normalization method, as it provides a guaranty that gradient directions are stable around the current point, allowing larger step widths.
        
        The \textbf{spectral norm} (or matrix norm) measures how far a matrix $\bm{A}$ can stretch a vector $\bm{x}$:
        $$||\bm{A}|| = \max_{\bm{x}\ne 0} \frac{||\bm{Ax}||}{||\bm{x}||}$$
        
        The numerical value of the spectral norm of $\bm{A}$ can be shown to be just its maximum singular value. To compute the maximum singular value, an algorithmic idea helps: the power iteration method, which yields the maximal eigenvector.
        
        \textbf{Power iteration} uses the fact that any matrix will rotate a random vector towards its largest eigenvector. Therefore, by iteratively calculating $\frac{\bm{AX}}{|\bm{Ax}|}$, the largest eigenvector is obtained eventually.
        
        In practice, it is observed that a single iteration is already sufficient to achieve the desired normalizing behavior.
    \end{nicebox}
\end{floatbox}

Consequently, the key ingredient is the Lipschitzness constraint of the discriminator network\footnote{The discriminator network in the context of continuous generator loss functions like the Wasserstein-based loss is called a ``critique'' network, as it no longer discriminates, but yields a metric. For ease of reading, this chapter sticks to the term ``discriminator''.}, and how to enforce this in a stable and regularized way. It soon turned out that weight clipping is not an ideal choice. Rather, two other methods have been proposed: the gradient penalty approach, and normalizing the weights with the spectral norm of the weight matrices.

Both have been added to the standard catalogue of performance boosting measures in GAN training ever since, where in particular spectral normalization (cf. \autoref{box:spectralnorm}) is attractive as it can be implemented very efficiently, has a sound theoretical and mathematical foundation, and ensures stable and efficient training.

\subsubsection{Example Application: Brain abnormality detection using WGAN}
\label{sec:wgan_application_example}

One of the first applications of Wasserstein GANs in a practical use case was presented in the medical domain, specifically in the context of attributing visible changes of a diseased patient with respect to a normal control to locations in the images~\cite{Baumgartner2018featureattribution}. The way this detection problem was cast into a GAN approach (and then solved with a Wasserstein GAN) was to delineate the regions that make the images of a diseased patient look ``diseased'', i.e. find the residual region, that, if subtracted from the diseased-looking image, would make it look ``normal''.

\begin{figure}[hbtp]
	\centering
		\includegraphics[width=0.95\textwidth]{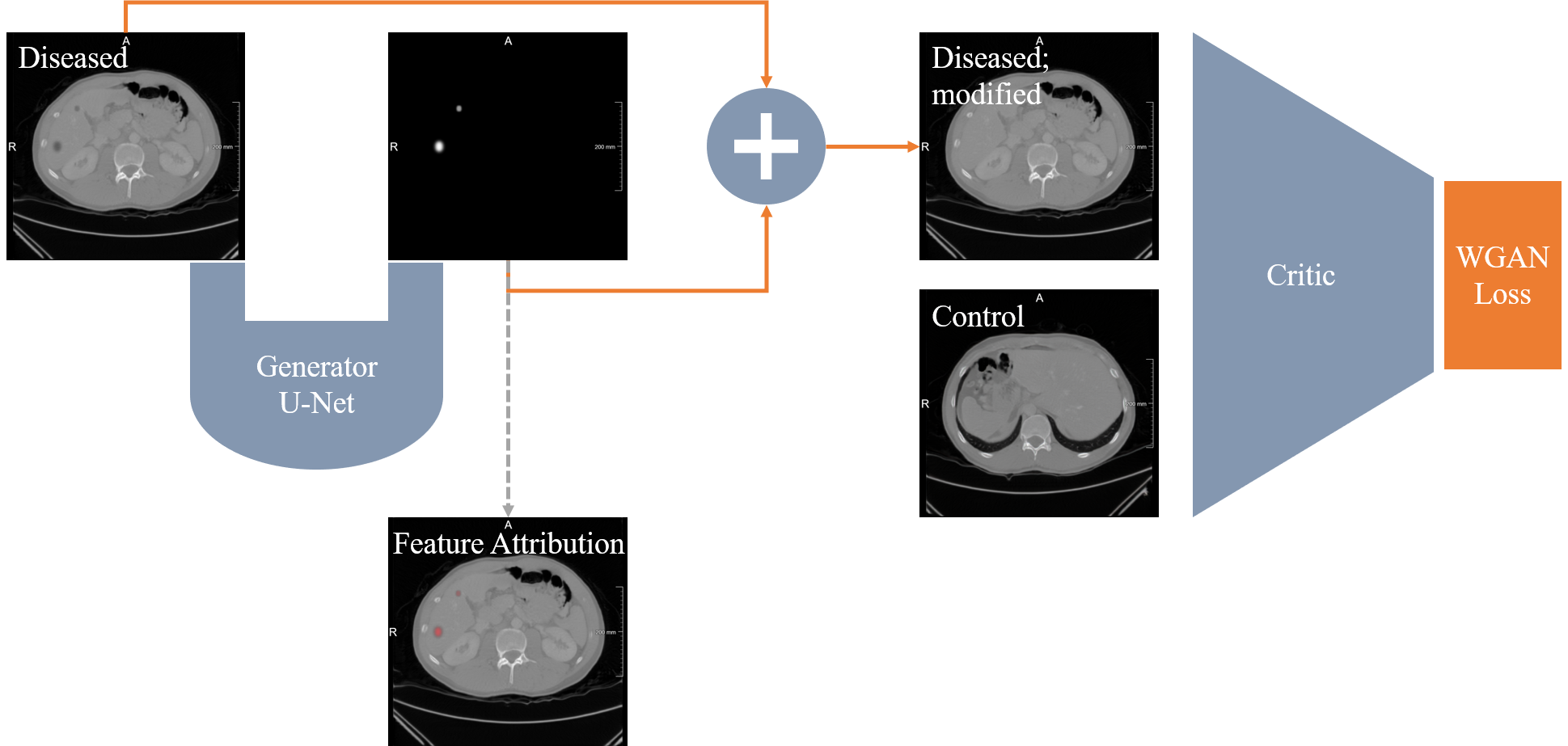}
	   \caption[VA-GAN Schematic]{An image of a diseased patient is run through a U-Net with the goal to yield a map that, if added to the input image, results in a modified image that fools the discriminator (``critique'') network into classifying it as a ``normal'' control. The map can be interpreted as the regions attributed to appear abnormal, giving rise to the name of the architecture: Visual Attribution GAN, VA-GAN.}
    \label{fig:VAGAN-schematic}
\end{figure}

\autoref{fig:VAGAN-schematic} shows the construction of the VA-GAN architecture with images from a mocked dataset for illustration. For the authors' results see their publication and code repository\footnote{\href{https://github.com/baumgach/vagan-code}{https://github.com/baumgach/vagan-code}}. 

For their implementation, the authors note that neither batch normalization nor layer normalization helped convergence, and hypothesize that the difference between real and generated examples may be a reason that in particular batch normalization may in fact have an adverse effect especially during the early training phase. Instead, they impose an $\ell_1$ norm loss component on the U-Net-generated ``visual (feature) attribution'' (VA) map to ensure it to be a minimal change to the subject. This serves to prevent the generator from changing the subject into some ``average normal'' image that it may otherwise learn. They employ an update regime that trains the critic network for more iterations than the generator, but doesn't train it to convergence as proposed in the original WGAN publications. Apart from these measures, in their code repository, the authors give several practical hints and heuristics that may stabilize the training, e.g. using a $tanh$ activation for the generator or exploring other dropout settings, and in general using a large enough dataset. They also point out that the Wasserstein distance isn't suited for model selection since it is too unstable and not directly correlated to the actual usefulness of the trained model.

This is one more reason to turn in the next section to an important topic in the context of validation for generative models: how to quantify their results?

\subsection{GAN performance metrics}
\label{ssec:perf_metrics}

One imminent question has so far been postponed, though it implicitly plays a crucial role in the quest for ``better'' GANs: how to actually measure the success of a GAN, or the performance in terms of result quality?

GANs can be adapted to solve image analysis tasks like segmentation or detection (cf. \autoref{sec:wgan_application_example}). In such cases, the quality and success can be measured in terms of task-related performance (Jaccard/Dice coefficient for segmentation, overlap metrics for detection etc.). 

Performance assessment is less trivial if the GAN is meant to generate unseen images from random vectors. In such scenarios, the intuitive criterion is how convincing the generated results are. But convincing to whom? One could expose human observers to the real and fake images, and ask them to tell them apart, and call a GAN better than a competing GAN if it fools the observer more consistently.\footnote{In fact, there is only very little research on the actual performance of GANs in fooling human observers, though guides exist on how to spot ``typical'' GAN artifacts in generated images. These are older than the latest GAN models, and it can be hypothesized that the lack of such literature is indirect confirmation of the overwhelming capacity of GANs to fool human observers.} Since this is practically infeasible, metrics were sought that provide a more objective assessment.

The most widely used way to assess GAN image quality is the Fr\'echet Inception Distance (FID). This distance is conceptually related to the Wasserstein distance. It has an analytical solution to calculate the distance of Gaussian (Normal) distributions. In the multivariate case, the Fr\'echet distance between two distributions $X$ and $Y$ is given by the squared distance of their means $\mu_X$ (resp. $\mu_Y$) and a term depending on the covariance matrix describing their variances $\Sigma_X$ (resp. $\Sigma_Y$):

\begin{equation}
    \label{eqn:frechetdistance}
    d(X,Y) = ||\mu_X - \mu_Y||^2 + \mathrm{Tr}(\Sigma_X+\Sigma_Y-2\sqrt{\Sigma_X\Sigma_Y}). 
\end{equation}

The way this distance function is being used is often the score, which is computed as follows: 
\begin{itemize}
    \item Take two batches of images (real/fake, respectively)
    \item Run them through a feature extraction or embedding model. For FID, the Inception model is used, pretrained on ImageNet. Retain the embeddings for all examples.
    \item Fit each one multivariate Normal distribution to the embedded real/fake examples.
    \item Calculate their Fr\'echet Distance according to the analytical formula in \autoref{eqn:frechetdistance}
\end{itemize}

This metric has a number of downsides. Typically, if computed for a larger batch of images, it decreases, although the same model is being evaluated. This bias can be remedied, but FID remains the most used metric still. Also, if the Inception network cannot capture the features of the data FID should be used on, it might simply be uninformative. This is obviously a grave concern in the medical domain where imaging features look much different from natural images (although, on the other hand, transfer learning for medical classification problems proved to work surprisingly well, so that apparently convolutional filters trained on photographs also extract applicable features from medical images). In any case, the selection of the pre-trained embedding model brings a bias into the validation results. Lastly, the assumption of a multivariate Normal distribution for the Inception features might not be accurate, and only describing it through their means and covariances is a severe reduction of information. Therefore, a qualitative evaluation is still required.

One obvious additional question arises: if the ultimate metric to judge the quality of the generator is given by, for example, the FID, why can't it be used as the optimization goal instead of minimizing a discriminator loss? In particular, as the Fr\'echet distance is a variant of the Wasserstein distance, an answer to this question is not obvious. In fact, feature matching as described in \autoref{box:stableGAN} exactly uses this type of idea, and likewise, it has been partially adopted in recent GAN architectures to enhance the stability of training with a more fine-grained loss component than a pure categorical cross-entropy loss on the ``real/fake'' classification of the discriminator. 

Related recent research is concerned with the question how generated results can automatically be detected to counteract fraudulent authors. So-called forensic algorithms detect patterns that point out generated images. This research puts up the question how to detect fake images reliably. Solutions based on different analysis directions encompass image fingerprinting and frequency-domain analysis~\cite{dzanic2020fourier,Joslin2020fakedetection,Le2021,Goebel2021}.

\section{Selected GAN Architectures You Should Know}
\label{sec:architectures}

In the following, we will examine some GAN architectures and GAN developments that were taken up by the medical community, or that address specific needs that might make them appealing e.g. for limited data scenarios. 

\subsection{Conditional GAN}
\label{sec:cGAN}
GANs cannot be told what to produce -- at least that was the case with early implementations. It was obvious, though, that a properly trained GAN would imprint the semantics of the domain onto its latent space, which was evidenced by experiments in which the latent space was traversed and images of certain characteristics could be produced by sampling accordingly. Also, it was found that certain dimensions of the latent space can correspond to certain features of the images, like hair color or glasses, so that modifying them alone can add or take away such visible traits. 

With the improved development of Conditional GANs~\cite{Isola2016cGAN} following a number of GANs that modeled the conditioning input more explicitly, another approach was introduced that was based on the U-Net architecture as a generator, and a favorable discriminator network that values local style over a full-image assessment.

Technically, the formulation of a conditional GAN is straight-forward. Recalling the value function (learning objective) of GANs from \autoref{eqn:gan-objective}:

\begin{equation*}
    J(G, D) = \mathbb{E}_{x \sim p_\mathrm{data}}[\log D(x)] + \mathbb{E}_{z \sim p_G}[1 - \log D(G(z))],
\end{equation*}

We now want to condition the generation on some additional knowledge or input. Consequently, both the generator $G$ and the discriminator $D$ will receive an additional ``conditioning'' input, which we call $x$. This can be a class label, but also any other associated information. Very commonly, the additional input will be an image, as for example for image translation application (e.g. transforming from one image modality to another such as for instance MRI to CT). The result is the cGAN objective function:

\begin{equation}
    \label{eqn:cgan-objective}
    J_\mathrm{cGAN}(G, D) = \mathbb{E}_{x \sim p_\mathrm{data}}[\log D(x\vert y)] + \mathbb{E}_{z \sim p_G}[1 - \log D(G(z \vert y))]
\end{equation}

\begin{figure}[hbtp]
	\centering
		\includegraphics[width=0.85\textwidth]{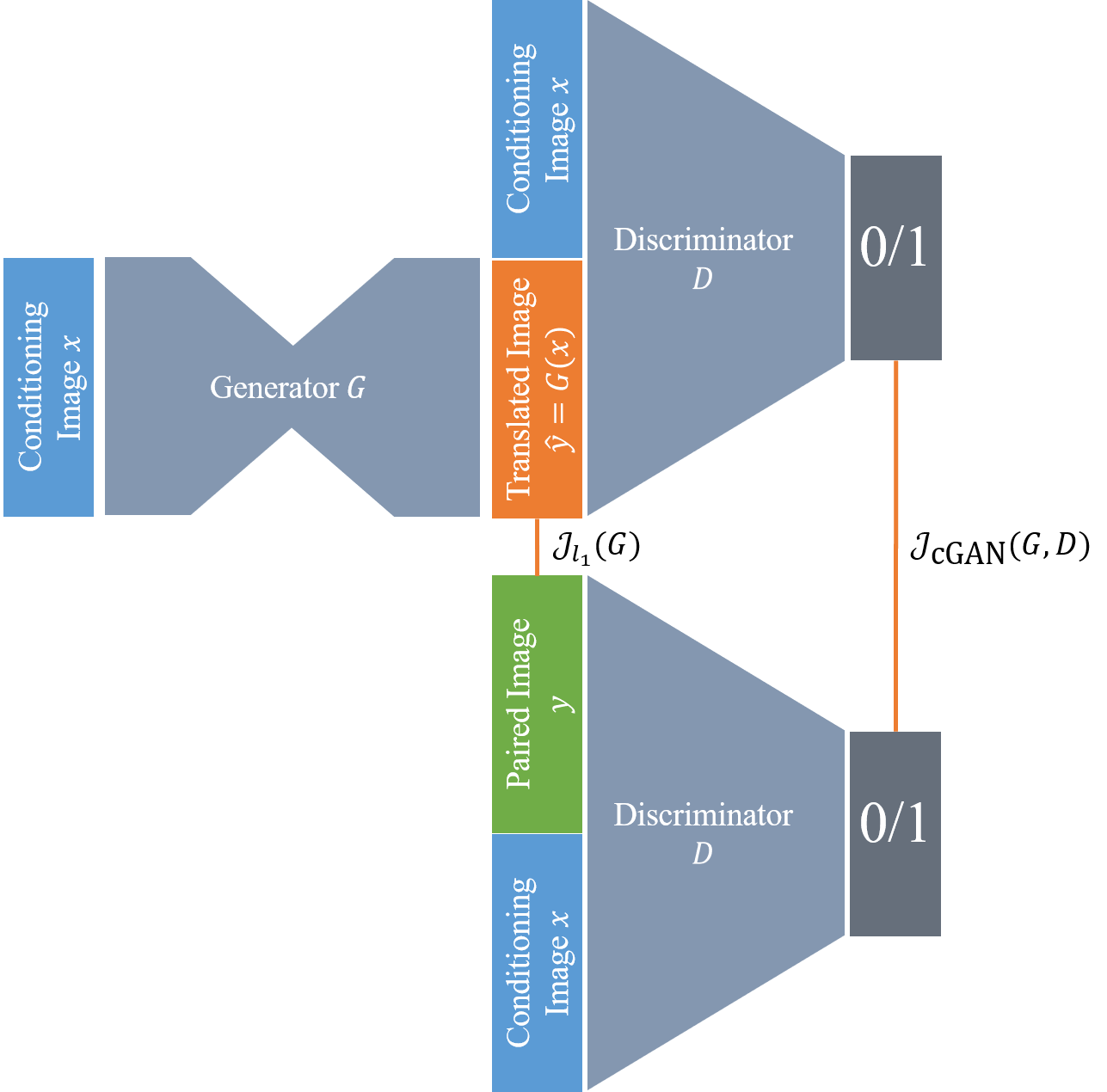}
	   \caption[cGAN Architecture]{A possible architecture for a cGAN. Left: The generator network takes the base images $x$ as input, and generates a translated image $\hat{y}$. The discriminator receives either this pair of images or a true pair $x,y$ (right). The additional generator reconstruction loss (often a $\ell_1$ loss) is calculated between $y$ and $\hat{y}$.}
    \label{fig:cGAN-arch}
\end{figure}

\citeauthor{Isola2016cGAN}~\cite{Isola2016cGAN} describe experiments with MNIST handwritten digits, where a simple generator with two layers of fully connected neurons was used, and similarly for the discriminator. $x$ was set to be the class label. In a second experiment, a CNN creates a feature representation of images, and the generator is trained to generate textual labels (choosing from a vocabulary of about 250.000 encoded terms) for the images conditioned on this feature representation.

\autoref{fig:cGAN-arch} shows a possible architecture to employ a cGAN architecture for image-to-image translation. In this diagram, the conditioning input is the target image that the trained network shall be able to produce based on some image input. The generator network therefore is a U-Net. The discriminator network can be implemented for example by a classification network. This network always receives two inputs: the conditioning image ($x$ in \autoref{fig:cGAN-arch}), and either the generated output $\hat{y}$ or the true paired image $y$.  

Note that the work of \citeauthor{Isola2016cGAN} introduces an additional loss term on the generator that measures the $\ell_1$ distance between the generated and ground truth image, which is (with variables as in \autoref{eqn:cgan-objective})

\begin{equation*}
    \mathcal{J}_{\ell_1}(G) = \mathbb{E}_{\textnormal{x,y,z}}\Vert y-G(x,z) \Vert _1,
\end{equation*}

where $\Vert \cdot \Vert_1$ is the $\ell_1$ norm.

The authors do not further justify this loss term apart from stating that $\ell_1$ is preferred over $\ell_2$ to encourage less blurry results. It can be expected that this loss component provides a good training signal to the generator when the discriminator loss doesn't, e.g. in the beginning of the training with little or no overlap of target and parameterized distributions. The authors propose to give the $\ell_1$ loss orders of magnitudes more weight than the discriminator loss component to value accurate translations of images over ``just'' very plausible images in the target domain.

\begin{figure}[hbtp]
	\centering
		\includegraphics[width=0.75\textwidth]{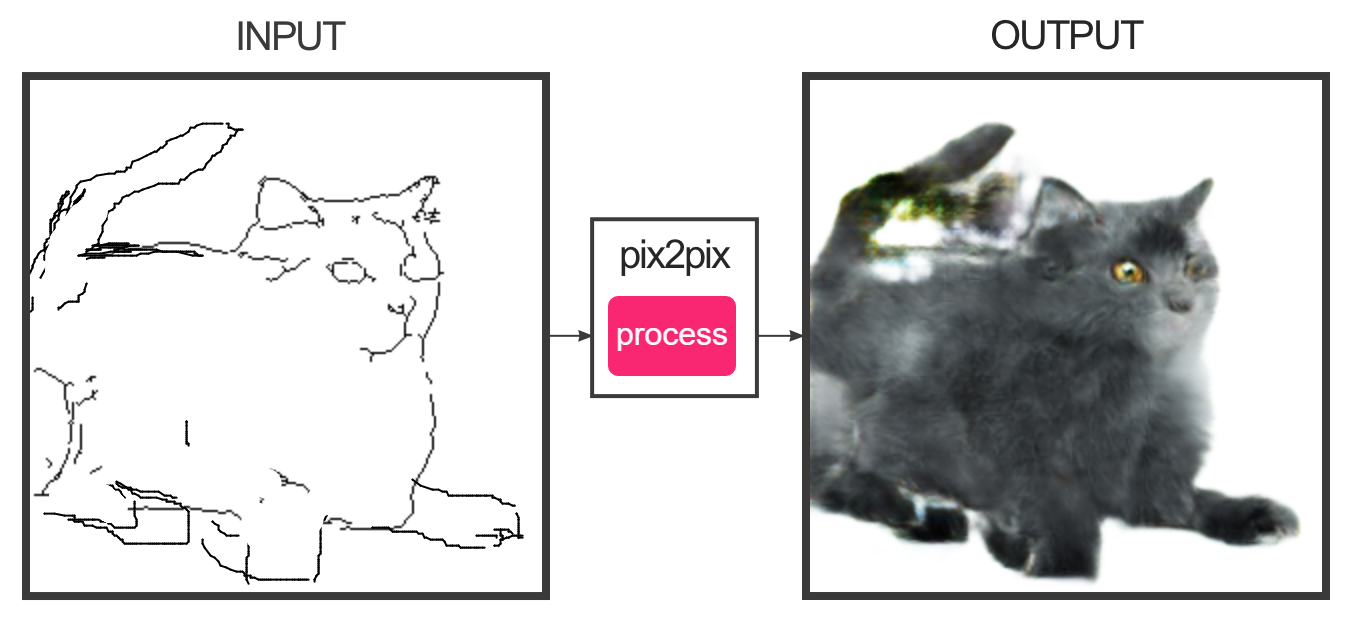}
	   \caption[pix2pix result]{Input and output of a pix2pix experiment. Online demo at \url{https://affinelayer.com/pixsrv/}}
    \label{fig:pix2pix}
\end{figure}

The cGAN, namely in the configuration with a U-Net serving as the generative network, was very quickly adopted by artists and scientists, thanks to the free implementation pix2pix\footnote{\url{https://github.com/phillipi/pix2pix}}. One example created with pix2pix is given in \autoref{fig:pix2pix}, where the cGAN was trained to produce cat images from line drawings.

One application in the medical domain was proposed for example by \citeauthor{Senaras2018cGANhisto}~\cite{Senaras2018cGANhisto}. The authors used a U-Net as a generator to produce a stained histopathology image from a label image that has two distinct labels for two kinds of cell nuclei. Here, the label image is the conditioning input to the network. Consequently, the discriminator network, a classification CNN tailored to the patch-based classification of slides, receives two inputs: the histopathology image, and a label image. 

Another example employed an augmented version of the Conditional GAN to translate CT to MR images of the brain, including a localized uncertainty estimate about the image translation success. In this work, a Bayesian approach to model the uncertainty was taken by including dropout layers in the Generator model~\cite{zhao2021bayesian}.

Lastly, a 3D version of the pix2pix approach with a 3D U-Net as a generative network was devised to segment gliomas in multi-modal brain MRI using data from the 2020 The International Multimodal Brain Tumor Segmentation (BraTS) challenge~\cite{bakas2019identifying}). The authors called their derived model vox2vox, alluding to the extension to 3D data~\cite{cirillo2020vox2vox}.

More conditioning methods have been developed over the years, some of which will be sketched further on. It is common to this type of GANs that paired images are required to train the network.

\subsection{CycleGAN}
\label{sec:cycleGAN}
While cGANs require paired data for the gold standard and conditioning input, this is often hard to come by, in particular in medical use cases. Therefore, the development of the CycleGAN set a milestone as it alleviates this requirement and allows to train image-to-image translation networks without paired input samples.

\begin{figure}[hbtp]
	\centering
		\includegraphics[width=0.95\textwidth]{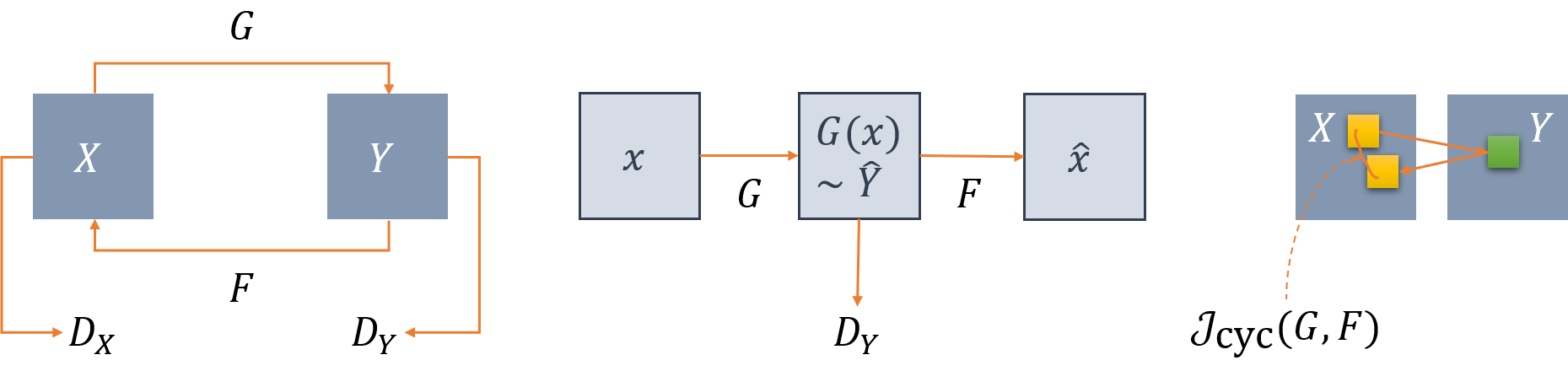}
	   \caption[Cycle GAN]{Cycle GAN. Left: Image translation functions $G$ and $F$ convert between two domains. Discriminators $D_X$ and $D_Y$ give adversarial losses in both domains. Middle: For one concrete translation of an image $x$, the translation to $Y$ and back to $X$ is depicted. Right: After the translation cycle, the original and back-translated result are compared in the \emph{cycle consistency loss}.}
    \label{fig:CycleGAN}
\end{figure}

The basic idea in this architecture is to train two mapping functions between two domains, and to execute them in sequence so that the resulting output is considered to be in the origin domain again. The output is compared against the original input, and their $\ell_1$ or $\ell_2$ distance establishes a novel addition to the otherwise usual adversarial GAN loss. This might conceptually remind one of the Autoencoder objectives: reproduce the input signal after encoding and decoding; only this time, there is no bottleneck, but another interpretable image space. This can be exploited to stabilize the training, since the sequential concatenation of image translation functions, which we will call $G$ and $F$, can be reversed. \autoref{fig:CycleGAN} shows a schematic of the overall process (left), and one incarnation of the cycle, here from image domain $X$ to $Y$ and back (middle). 

CycleGANs employ several loss terms in training: two adversarial losses $\mathcal{J}(G,D_Y)$ and $\mathcal{J}(\mathcal{F},D_X)$, and two cycle consistency losses, of which one, $\mathcal{J}_\mathrm{cyc}(G,\mathcal{F})$ is indicated rightmost in \autoref{fig:CycleGAN}. \citeauthor{Zhu2017b-cyclegan} presented the initial publication with a participation of the cGAN author Isola~\cite{Zhu2017b-cyclegan}. The cycle consistency losses are $\ell_1$ losses in their implementation, and the GAN losses are least square losses instead of negative log likelihood, since more stable training was observed with this choice.

\citeauthor{Almahairi2018-cyclegan-aug} provided an augmented version~\cite{Almahairi2018-cyclegan-aug}, noting that the original implementation suffers from the inability to generate stochastic results in the target domain $Y$, but rather learns a one-to-one mapping between $X$ and $Y$ and vice versa. To alleviate this problem, the generators are conditioned on one latent space each for both directions, so that, for the same input $x\in X$, $G$ will now produce multiple generated outputs in $Y$ depending on the sample from the auxiliary latent space (and similarly in reverse). Still, $\mathcal{F}$ has to re-create a $\hat{x}$ minimizing the cycle consistency loss for each of these samples. This also remedies a second criticism brought forward against vanilla CycleGANs: these networks can learn to hide information in the (intermediate) target image domain that fool the discriminator, but help the backward generator to minimize the cycle consistency loss more efficiently~\cite{Chu2017-cyclegan-stegano}. \citeauthor{Chu2017-cyclegan-stegano} use adaptive histogram equalization to show that in visually empty regions of the intermediate images information is present. This is a finding reminiscent of adversarial attacks, which the authors elaborate on in their publication.

\citeauthor{Zhang2018-cyclegan-medical}~\cite{Zhang2018-cyclegan-medical} show a medical application. In their work, a CycleGAN has been used to train image translation and segmentation models on unpaired images of the heart, acquired with MRI and CT, and with gold standard expert segmentations available for both imaging datasets. The authors proposed to learn more powerful segmentation models by enriching both datasets with artificially generated data. To this end, MRIs are converted into CT contrast images and vice versa using GANs. Segmentation models for MRI and CT are then trained on dataset consisting of original images and their expert segmentations, and augmented by the converted images, for which expert segmentations can be carried over from their original domain. To achieve this, it is of importance that the converted (translated) images accurately depict the shape of the organs as expected in the target domain, which is enforced using the shape consistency loss.

In the extended setup of the CycleGAN with shape and cycle consistency, three different loss types instead of the original two are combined during training:
\begin{description}
    \item[Adversarial GAN losses $\mathcal{J}_\mathrm{GAN}$.] This loss term is the same as defined e.g. in \autoref{eqn:gan-objective}.
    \item[Cycle consistency losses $\mathcal{J}_\mathrm{cyc}$.] This is the $\ell_1$ loss presented by the original CycleGAN authors discussed above. 
    \item[Shape consistency losses $\mathcal{J}_\mathrm{shape}$.] The shape consistency loss is a new addition proposed by the authors. A cross correlation loss takes into account two segmentations, the first being the gold standard segmentation $m_x$ for an $x\in X$, and one segmentation produced by a Segmenter network $S$ that was trained on domain $Y$ and receives the translated image $\hat{y} = G(x)$. 
\end{description}

\begin{figure}[hbtp]
	\centering
		\includegraphics[width=0.95\textwidth]{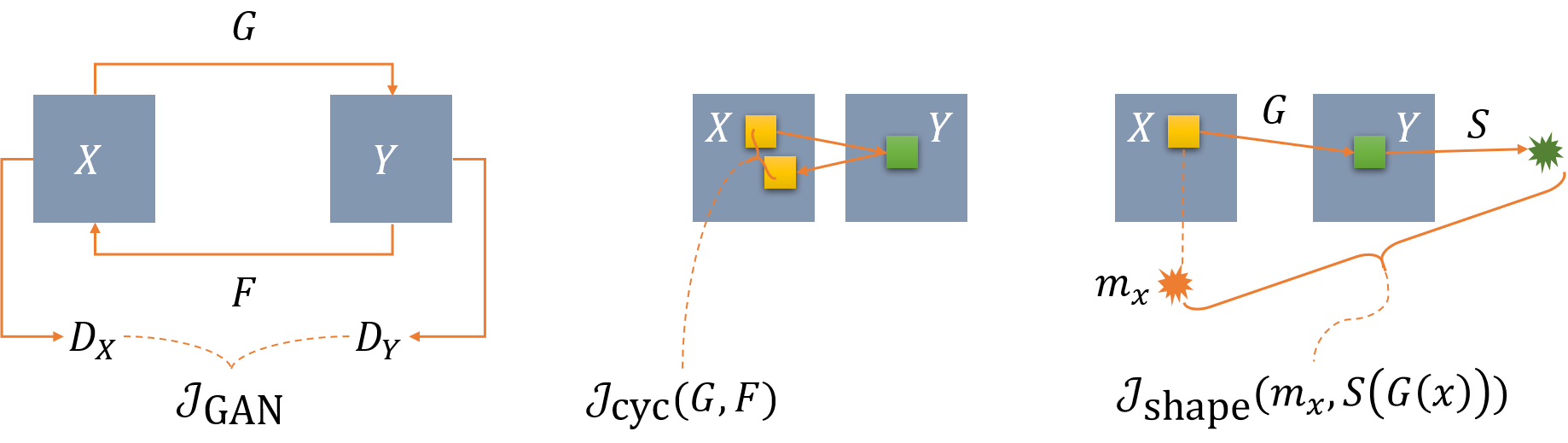}
	   \caption[CycleGAN with shape consistency loss]{Cycle GAN with shape consistency loss (rightmost part of figure). Note that the Figure shows only one direction to ease readability.}
    \label{fig:CycleGAN-Shape}
\end{figure}

\autoref{fig:CycleGAN-Shape} depicts the three loss components, of which the first two are known already from \autoref{fig:CycleGAN}.

Note that the description as well as \autoref{fig:CycleGAN-Shape} only show one direction for cycle and shape consistency loss. Both are duplicated into the other direction, and combined into the overall training objective, which then consists of six components.

In several other works, the CycleGAN approach was extended and combined with domain adaption methods for various segmentation tasks, and also extended to volumetric data~\cite{hoffman2017cycada,huo2018adversarialsynth,Yang2020}.

\subsection{StyleGAN and Successor}
\label{sec:styleGAN}
One of the most powerful image synthesis GANs to date is the successor of StyleGAN, StyleGAN2~\cite{Karras2018stylegan,Karras2020stylegan2}. The authors, at the time of writing researching at Nvidia, deviate from the usual GAN approach in which an image is generated from a randomly sampled vector from a latent space. Instead, they use a latent space that is created by a mapping function $f$ which is in their architecture implemented as a multilayer perceptron which maps from a 512-dimensional space $Z$ into a 512-dimensional space $W$. The second major change consisted of the so-called adaptive instance normalization layer, AdaIN, which implements a normalisation to zero-mean and unit variance of each feature map, followed by a multiplicative factor and an additive bias term. This serves to re-weight the importance of feature maps in one layer. To ensure the locality of the re-weighting, the operation is followed by the non-linearity. The scaling and bias are two components of $\mathbf{y}=(y_s,y_b)$, which is the result of a learnable affine transformation $A$ applied to a sample from $W$. 

In their experiments, \citeauthor{Karras2018stylegan}~\cite{Karras2018stylegan} recognized that after these changes, the GAN actually no longer depended on the input vector drawn from $W$ itself, so the random latent vector was replaced by a static vector fed into the GAN. The $\mathbf{y}$, which they call \emph{styles}, remained to be results from a vector randomly sampled from the new embedding space $W$.

\begin{figure}[hbtp]
	\centering
		\includegraphics[width=1.0\textwidth]{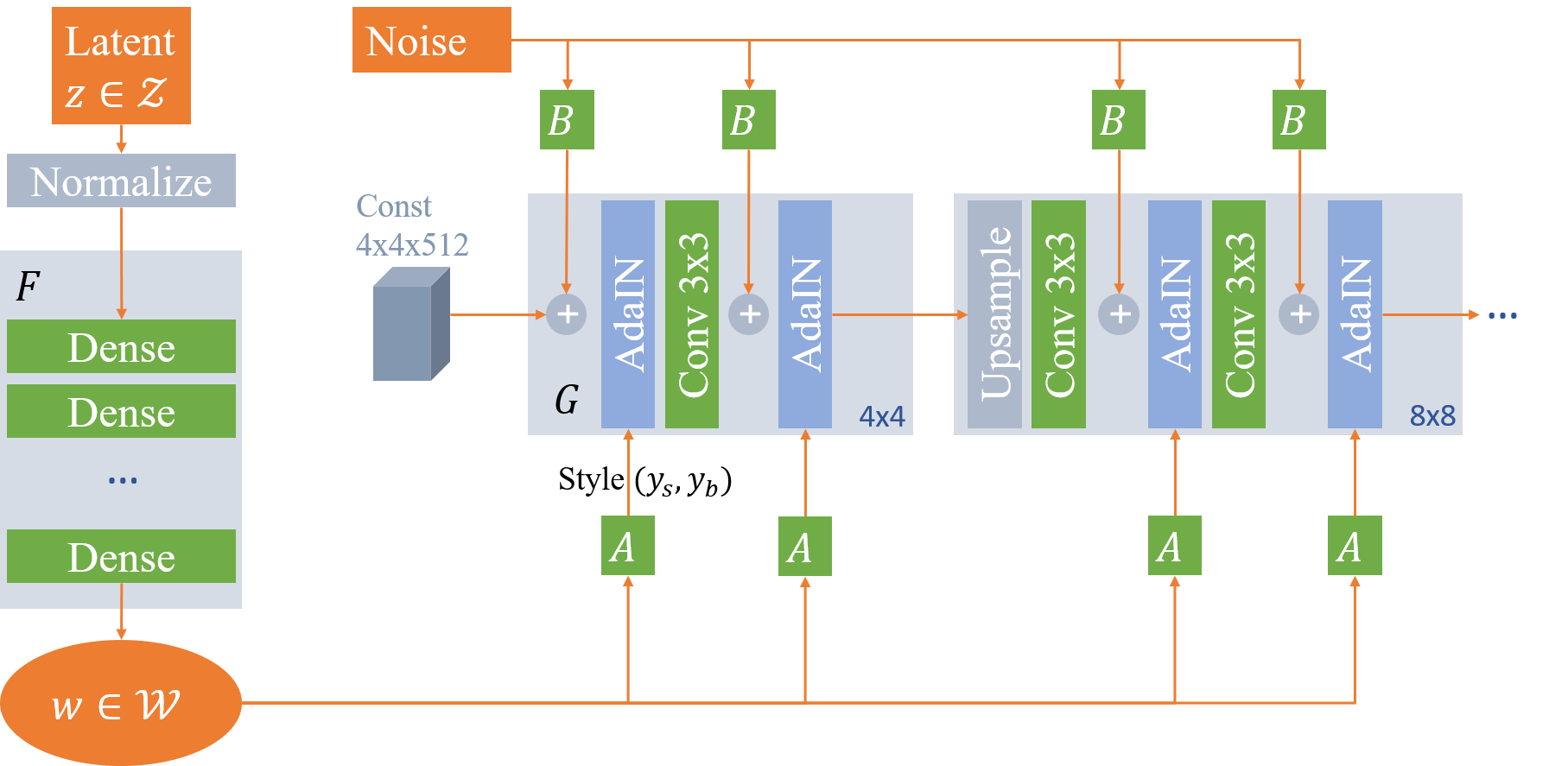}
	   \caption[StyleGAN Architecture]{StyleGAN architecture, after \cite{Karras2018stylegan}. Learnable layers and transformations are shown in green, the AdaIN function in blue.}
    \label{fig:StyleGAN-arch}
\end{figure}

Lastly, noise is added in each layer, which serves to allow the GAN to produce more variation without learning to produce it from actual image content. The noise, like the latent vector, is fed through learnable transformations $B$, before it is added to the unnormalized feature maps. The overall architecture is sketched in \autoref{fig:StyleGAN-arch}.

\begin{figure}[hbtp]
	\centering
		\includegraphics[width=0.98\textwidth]{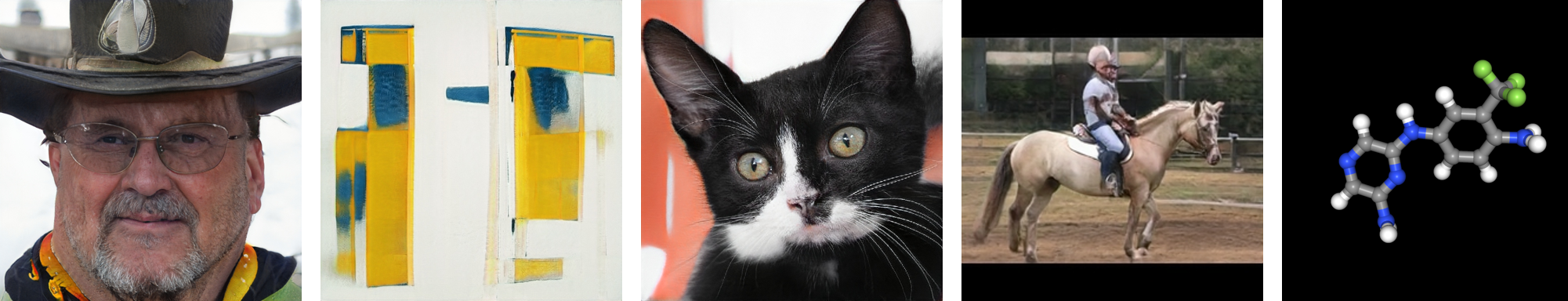}
	   \caption[StyleGAN results]{Images created with StyleGAN; {\url{https://this{person|artwork|cat|horse|chemical}doesnotexist.com}. Last accessed: 2022-01-14}}
    \label{fig:styleganresults}
\end{figure}

In the basic setup, one sample is drawn from $W$, and fed through per-layer learned $A$ to gain per-layer different interpretations of the style, $\mathbf{y}=(y_s, y_b)$. This can be changed, however, and the authors show how using one random sample $w_1$ in some of the layer blocks, and another sample $w_2$ in the remaining, the result will be a mixture of styles of both individual samples. This way, the coarse attributes of the generated image can stem from one sample, and the fine detail from another. Applied to a face generator, for example pose and shape of the face are determined in the coarse early layers of the network, while hair structure and skin texture are the fine details of the last layers. The architecture and results gained widespread attention through a website\footnote{\url{https://thispersondoesnotexist.com/}}, which recently was followed up by further similar pages. Results are depicted in
\autoref{fig:styleganresults}.

The crucial finding in StyleGAN was that the mapping function $F$ transforming the latent space vector from $\mathcal{Z}$ to $\mathcal{W}$ serves to ensure a disentangled (flattened) latent space. Practically, this means that if interpolating points $z_i$ between two points $z_1$, $z_2$ drawn from $\mathcal{Z}$, and reconstructing images from these interpolated points $z_i$, semantic objects might appear (in a StyleGAN generating faces, for example a hat or glasses) that are neither part of the generated images from the first point $z_1$, nor the second point $z_2$ between which it has been interpolated. Conversely, if interpolating in $\mathcal{W}$, this ``semantic discontinuity'' is no longer the case, as the authors show with experiments in which they measure the visual change of resulting images when traversing both latent spaces.

In their follow-up publications, the same authors improve the performance even further. They stick to the basic architecture, but redesign the generative network pertaining to the AdaIN function. In addition, they add their metric from~\cite{Karras2018stylegan} that was meant to quantify the entanglement of the latent space as a regularizer. The discriminator network was also enhanced, and the mechanisms of StyleGAN that implement the progressive growing have been successively replaced by more performance-efficient setups. In their experiments, they show a growth of visual and measured quality and removal of several artifacts reported for StyleGAN~\cite{Karras2020stylegan2}.


\subsection{Stabilized GAN for Few-Shot Learning}
\label{sec:stabilizedGAN}
GAN training was very demanding both regarding GPU power, in particular for high-performance architectures like StyleGAN and StyleGAN2, and, as importantly, availability of data. StyleGAN2, for example, has typical training times of about 10 days on a Nvidia 8-GPU Tesla V100. The datasets comprised at least 10s of thousands of images, and easily orders of magnitude more. Particularly in the medical domain, such richness of data is typically hard to find.

The authors of ~\cite{liu2021stabilizedGAN}  propose simple measures to stabilize the training of a specific GAN architecture, which they design from scratch using a replacement for residual blocks, arranged in an architecture with very few convolutional layers, and a loss that drives the discriminator to be less certain when it gets closer to convergence. In sum, this achieves very fast training and yields results competitive with prior GANs~\cite{liu2021stabilizedGAN} and outperforming them in low-data situations.

The key ingredients to the architecture are shortcut connections in the generator model that rescale feature maps of higher resolution with learnable weights derived from low resolutions. The effect is to make fine details simultaneously more independent of direct predecessor feature maps, and yet ensure consistency across scales.

\begin{figure}[hbtp]
	\centering
		\includegraphics[width=0.98\textwidth]{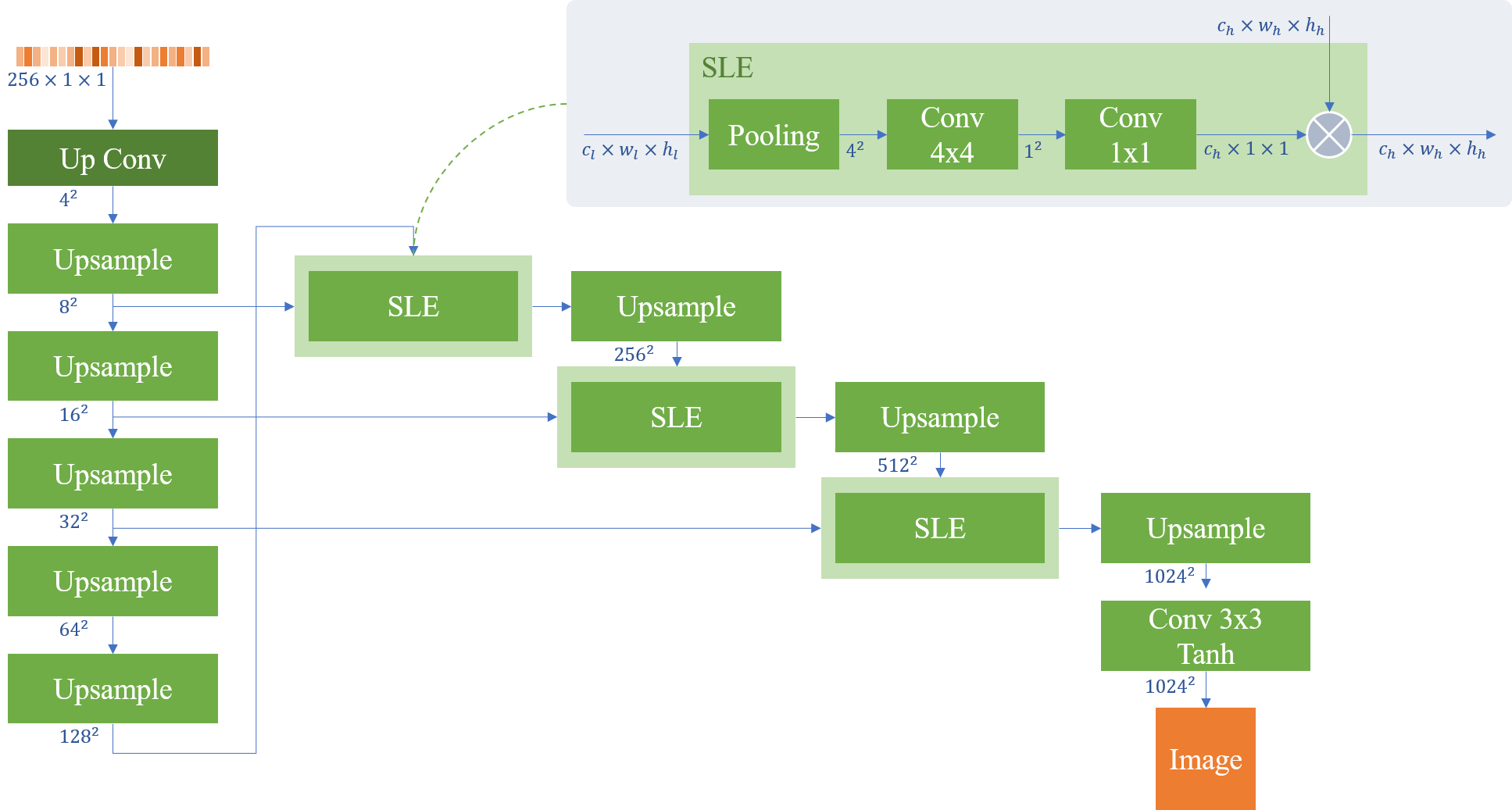}
	   \caption[FastGAN Generator]{The FastGAN generator network. Shortcut connections through feature map weighting layers (called skip-layer excitation, SLE) transport information from low-resolution feature maps into high-resolution feature maps. For details regarding the blocks, see text.}
    \label{fig:fastgangen}
\end{figure}

A random seed vector of length 256 enters the first block (``Up Conv''), where it is upscaled to a $256 \times 4 \times 4$ tensor. In \autoref{fig:fastgangen}, the further key blocks of the architecture are ``Upsample'' and ``SLE'' blocks.

\begin{description}
    \item[Upsample] blocks consist of a nearest-neighbor upsampling followed by a $3\times 3$ convolution, batch normalisation and nonlinearity.
    \item[SLE] blocks (seen in the top right inset in the architecture diagram) don't touch the incoming high-resolution input (entering from top into the block), but comprise a pooling layer that in each SLE block is set up to yield a $4 \times 4$ stack of feature maps, followed by a convolution to reduce to a $1\times 1$ tensor, that is then in a $1 \times 1$ convolution brought to the same number of channels as the high-resolution input. This vector is then multiplied to the channels of the high-resolution input.
\end{description}

\begin{figure}[hbtp]
	\centering
		\includegraphics[width=0.98\textwidth]{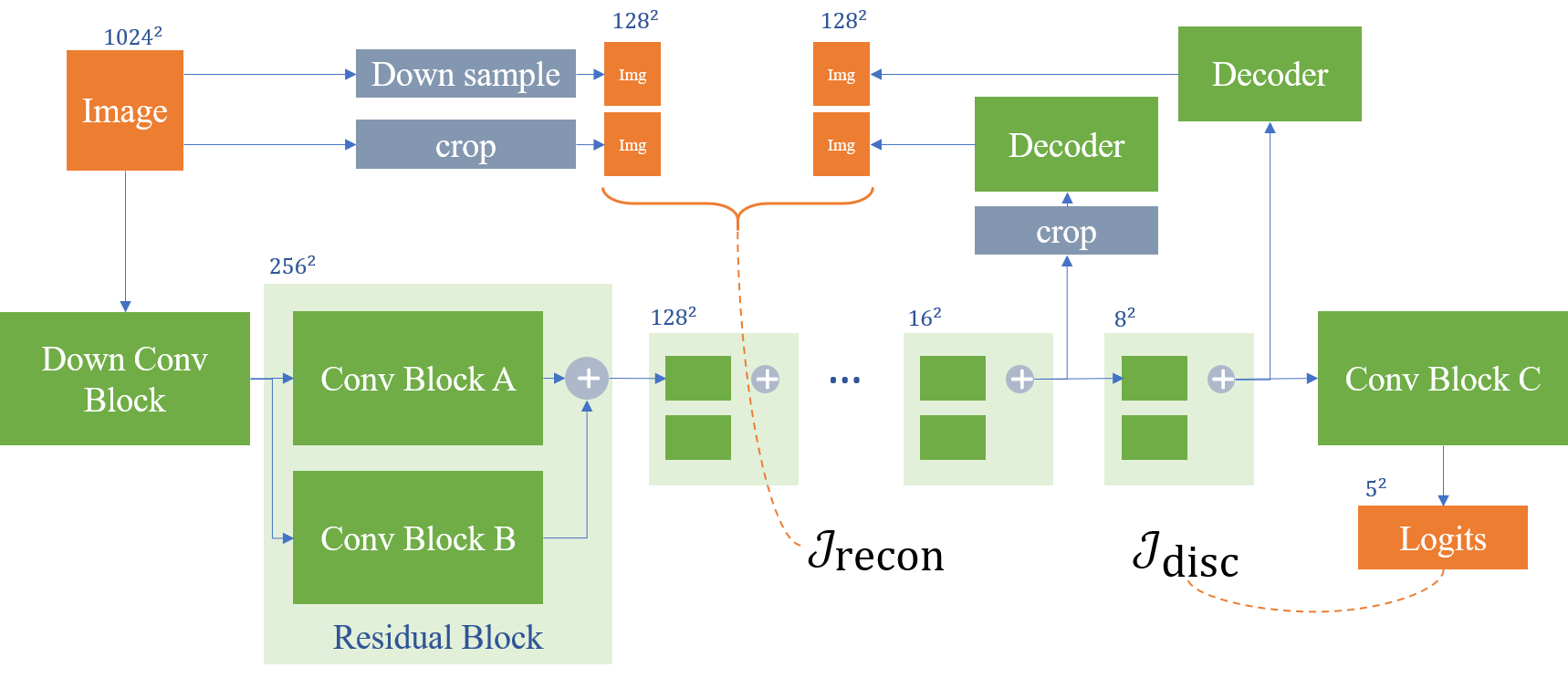}
	   \caption[FastGAN Discriminator]{The FastGAN self-supervision mechanism of the discriminator network. Self-supervision manifests through the loss term indicated by the curly bracket between reconstructions from feature maps and resampled/cropped versions of the original real image, $\mathcal{J}_\mathrm{recon}$.}
    \label{fig:fastganssd}
\end{figure}

Secondly, the architecture introduces a self-supervision feature in the discriminator network. The discriminator network (see \autoref{fig:fastganssd}) is a simple CNN with strided convolutions in each layer, halving resolution in each feature map. In the latest (coarsest) feature maps, simple up-scaling convolutional networks are attached that generate small images, which are then compared in loss functions ($\mathcal{J}_\mathrm{recon}$ in \autoref{fig:fastganssd}) to down-sampled versions of the real input image. This self-supervision of the discriminator is only performed for real images, not for generated ones.

The blocks in the figure spell out as follows:
\begin{description}
    \item[Down Conv Block] consists of two Convolutional layers with strided 4x4 convolutions, effectively reducing the resolution from $1024^2$ to $256^2$.
    \item[Residual Blocks] have two sub-items, ``Conv Block A'' being a strided $4 \times 4$ convolution to half resolution, followed by a padded $3 \times 3$ convolution. ``Conv Block B'' consists of a strided $2 \times 2$ average pooling that quarters resolution, followed by a $1 \times 1$ convolution, so that both blocks result in identically shaped tensors, which are then added.
    \item[Conv Block C] consists of a $1 \times 1$ convolution followed by a $4 \times 4$ convolution without strides or padding, so that the incoming $8^2$ feature map is reduced to $5^2$.
    \item[Decoder] The Decoder networks are four blocks of up-sampling layers each followed by 3x3 convolutions.
\end{description}

The losses employed in the model are the discriminator loss consisting of the hinge version of the usual GAN loss, with the added regularizing reconstruction loss between original real samples and their reconstruction, and the generator loss plainly being $\mathcal{J}_\mathrm{G} = \mathbb{E}_{z \sim Z}[D(G(z))]$. 

\begin{figure}[htbp]
	\centering
		\includegraphics[width=1.0\textwidth]{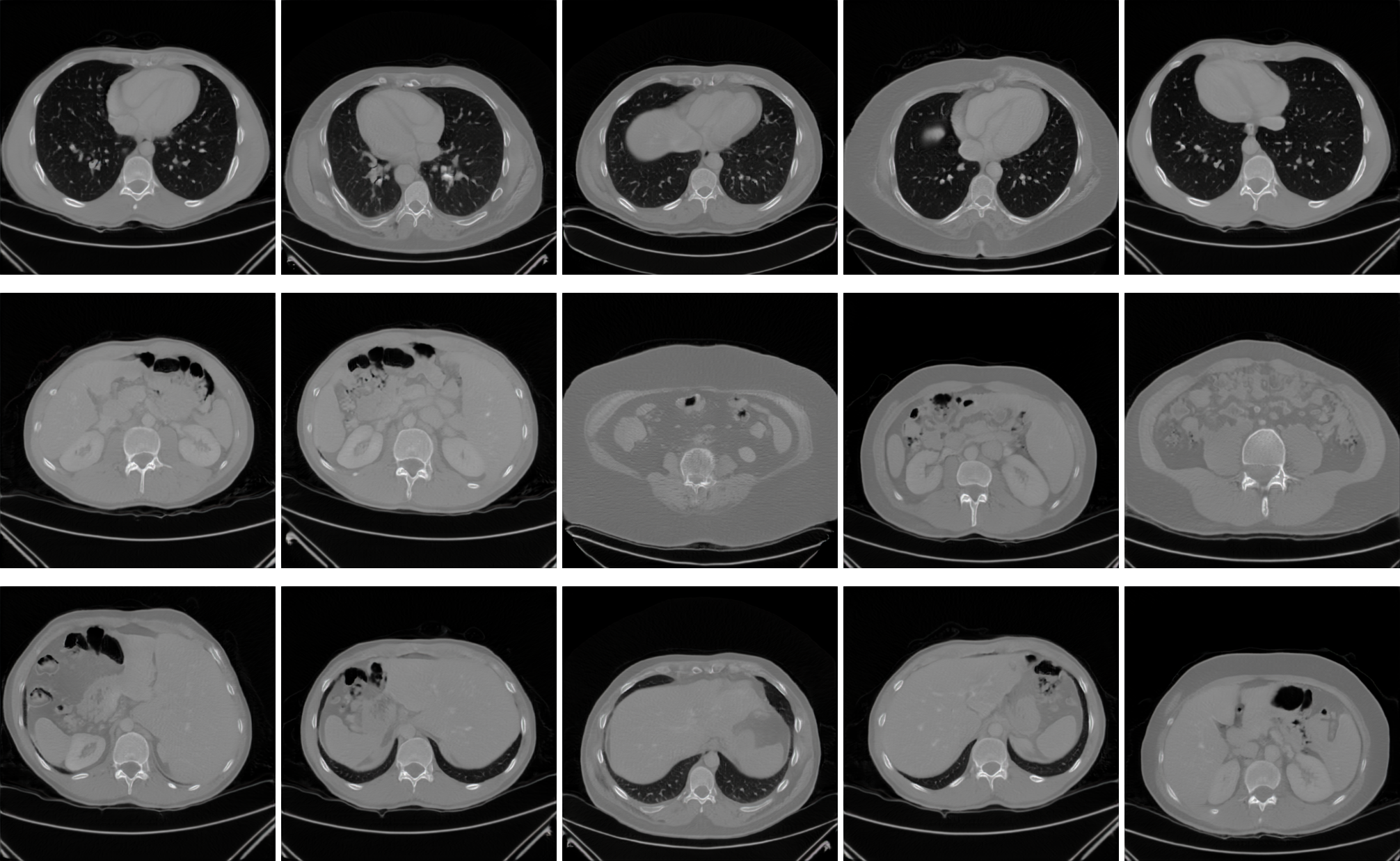}
	   \caption[Example FastGAN generated CT slices]{FastGAN as implemented by the authors has been used to train a CT slice generative model. Images are not cherry-picked, but arranged by similar anatomical regions.}
	\label{fig:fastgan_result}
\end{figure}

The model is easy to train on modest hardware and little data, as evidenced by own experiments on a set of about 30 chest CTs (about 2500 image slices, converted to RGB). \autoref{fig:fastgan_result} shows randomly picked generated example slices, roughly arranged by anatomical content. It is to be noted that organs appear mirrored in some images. On the other hand, no color artifacts are visible, so that the model has learned to produce only grey scale images. Training time for 50,000 iterations on a Nvidia TitanX GPU was approximately 10 hours.

\subsection{VQGAN}
\label{sec:vqgan}

In a recent development, a team of researchers combined techniques for text interpretation with a dictionary of elementary image elements feeding into a generative network. The basic architecture component that is employed goes back to vector quantization variational autoencoders (VQ-VAE), where the latent space is no longer allowed to be continuous, but is quantized. This allows to use the latent space vectors in a look-up table: the visual elements.

\begin{figure}[htbp]
	\centering
		\includegraphics[width=0.5\textwidth]{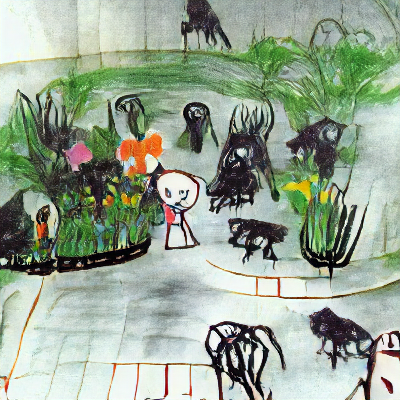}
	   \caption[Example VQGAN generated image]{The VQGAN+CLIP combination creates images from text inputs, here: ``A child drawing of a dark garden full of animals''.}
	\label{fig:vqgan-example}
\end{figure}

\autoref{fig:vqgan-example} was created using code available \href{https://colab.research.google.com/drive/1ZAus_gn2RhTZWzOWUpPERNC0Q8OhZRTZ}{online}, that demonstrates how images of different visual styles can be created using the combination of text-based conditioning and a powerful generative network. 

The basis for image generation is the VQGAN (``Vector Quantization Generative Adversarial Network'')~\cite{esser2021taming}, which learns representations of input images that can later steer the generative process, in an adversarial framework. The conditioning is achieved with the CLIP (``Contrastive Image-Language Pretraining'') model that learns a discriminator that can judge plausible images for a text label or vice versa~\cite{radford2021clip-learning}. 

The architecture has been developed with an observation in mind that puts the benefits and drawbacks of convolutional and transformer architectures in relation to each other. While the locality bias of convolutional architectures is inappropriate if overall structural image relations should be considered, it is of great help in capturing textural details that can exist anywhere, like fur, hair, pavement or grass, but where the exact representation of hair positions or pavement stones is irrelevant. On the other hand, image transformers are known to learn convolutional operators implicitly, posing a severe computational burden without a visible impact on the results. Therefore, \citeauthor{esser2021taming}~\cite{esser2021taming} suggest to combine convolutional operators for local detail representation, and transformer-based components for image structure.

Since the VQGAN as a whole is no longer a pure CNN, but for a crucial component uses a transformer architecture, this model will be brought up again briefly in \autoref{sec:transformer_models}. 

The VQGAN architecture is derived from the VQ-VAE, the Vector Quantization Variational Autoencoder~\cite{Oord2017vqvae}, adding a reconstruction loss through a discriminator, which turns it into a GAN. At the core of the architecture is the quantization of estimated codebook entries. Among the quantised entries in the codebook, the closest entry to the query vector coding an image patch is determined. The found codebook entry is then referred to by its index in the codebook. This quantization operation is non-differentiable, so for end-to-end training, gradients are simply copied through it during back-propagation.  

The transformer can then efficiently learn to predict codebook indices from those comprising the current version of the image, and the generative part of the architecture, the decoder, produces a new version of the image. Learning expressive codebook entries is enforced by a perceptual loss that punishes inaccurate local texture, etc. Through this, the authors can show that high compression levels can be achieved -- a prerequisite to enable efficient, yet comprehensive transformer training.


\section{Other Generative Models}
\label{sec:other_gen_models}

We have already seen how GANs were not the first approach to image generation, but have prevailed for a time when they became computationally feasible and in consequence have been better understood and improved to accomplish tasks in image analysis and image generation with great success. In parallel with GANs, other fundamentally different generative modeling approaches have also been under continued development, most of which have precursors from the ``before-GAN'' era as well. To give a comprehensive outlook, we will sketch in this last section the state-of-the-art of a selection of these approaches.\footnote{The research on so-called flow-based models, e.g. Normalizing Flows, has been omitted in this chapter, though acknowledging their emerging relevance also in the context of image generation. Flow-based models are built from sequences of invertible transformations, so that they learn data distributions explicitly at the expense of sometimes higher computational costs due to their sequential architecture. When combined e.g. with a powerful GAN, they allow innovative applications, for example to steer the exploration of a GAN's latent space to achieve fine-grained control over semantic attributes for conditional image generation. Interested readers are referred to the literature~\cite{JimenezRezende2015VariationalIW,Dinh2017RealNVP,weng2018flow,kingma2018Glow,Abdal_2021}.}

\subsection{Diffusion and Score based Models}
\label{sec:diffusion}

Diffusion models take a completely different approach to distribution estimation. GANs implicitly represent the target distribution by learning a surrogate distribution. Likelihood-based models like VAE approximate the target distribution explicitly, not requiring the surrogate. In diffusion models, however, the gradient of the log probability density function is estimated, instead of looking at the distribution itself (which would be the unfathomable integral of the gradient). This value is known as the Stein score function, leading to the notion that diffusion models are one variant of score-based models~\cite{song2021scorebased}. 

\begin{figure}[hbtp]
	\centering
		\includegraphics[width=0.95\textwidth]{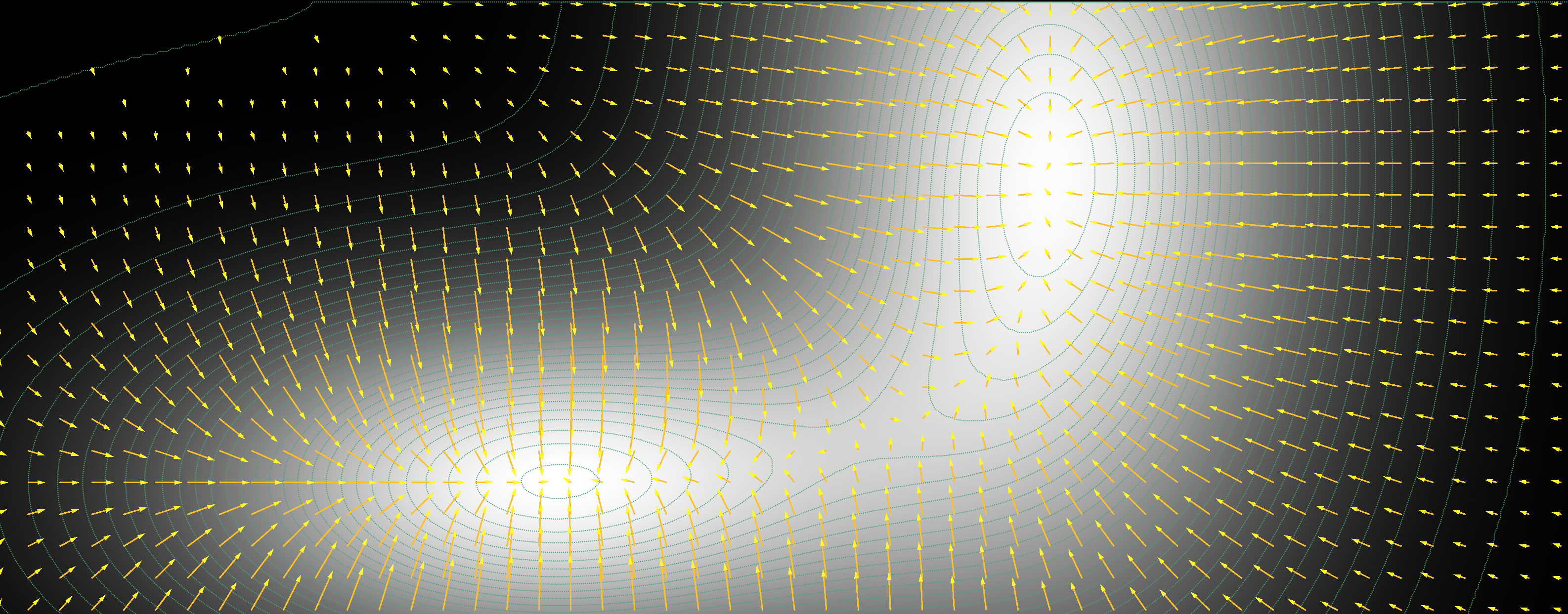}
	   \caption[Stein score function illustration]{The Stein score function can be conceived of as the gradient of the log probability density function, here indicated by two Gaussians. The arrows represent the score function.}
	\label{fig:scorefunction}
\end{figure}

The simple idea behind this class of models is to revert a sequential noising process. Consider some image. Then, perform a large number of steps. In each step, add a small amount of noise from a known distribution, e.g. the Normal distribution. Do this until the result is indistinguishable from random noise.  

The denoising process is then formulated as a latent variable model, where $T-1$ latents successively progress from a noise image $\mathbf{x}_T \sim \mathcal{N}(\mathbf{x}_T;\mathbf{0},\mathbf{I})$ to the reconstruction that we call $\mathbf{x}_0 \sim q(\mathbf{x}_0)$. The reconstructed image, $\mathbf{x_0}$, is therefore obtained by a \emph{reverse process} $q_\theta(\mathbf{x}_{0:T})$. Note that each step in this chain can be evaluated in closed form~\cite{ho2020denoising}. Several model implementations of this approach exist, one being the Deep Diffusion Probabilistic Model, DDPM. Here, a deep neural network learns to perform one denoising step given the so-far achieved image and a $t \in \{1, \ldots, T \}$. Iterative application of the model to the result of the last iteration will eventually yield a generated image from noise input. 

Autoregressive Diffusion Models (ARDMs)~\cite{hoogeboom2021autoregressive} follow yet another thought model, roughly reminiscent of PixelRNNs we have briefly mentioned above (see \autoref{sec:vae}). Both share the approach to condition the prediction of the next pixel or pixels on the already predicted ones. Other than in the PixelRNN, however, the specific ARDM proposed by the authors does not rely on a predetermined schedule of pixel updates, so that these models can be categorized as latent variable models.

As of late, the general topic of score-based methods, among which diffusion models are one variant, received more attention in the research community, fueled by a growing body of publications that report image synthesis results that outperform GANs~\cite{song2021scorebased,Dhariwal2021DDPMbeatGAN,Nichol2021improved_DDPM}. Score function-based and diffusion models superficially share the similar concept of sequentially adding/removing noise, but achieve their objective with very different means: where score function-based approaches are trained by score matching, and their sampling process uses Langevin dynamics \cite{song2019generative}, diffusion models are trained using the evidence lower bound (ELBO) and sample with a decoder, which is commonly a neural network.

Score function-based (sometimes also: score matching) generative models have been developed to astounding quality levels, and the recent works of Yang Song and others provide accessible blog posts\footnote{\url{https://yang-song.github.io/blog/}}, and a comprehensive treatment of the subject in several publications~\cite{song2019generative,song2019sliced,song2021scorebased}.

In the work of~\citeauthor{ho2020denoising}~\cite{ho2020denoising}, the step-wise reverse (denoising) process is the basis of the Denoising Diffusion Probabilistic Models (DDPM). The authors emphasize that a proper selection of the noise schedule is crucial to fast, yet high-quality, results. They point out, that their work is a combination of diffusion probabilistic models with score-matching models, in this combination also generalizing and including the ideas of autoregressive denoising models. In an extension of \citeauthor{ho2020denoising}'s work by \citeauthor{Nichol2021improved_DDPM}~\cite{Nichol2021improved_DDPM}, an importance sampling scheme was introduced that lets the denoising process steer the most easy to predict next image elements. Equipped with this new addition, the authors can show that, in comparison to GANs, a wider region of the target distribution is covered by the generative model. 


\subsection{Transformer-based Generative Models}
\label{sec:transformer_models}

The basics of how attention mechanisms and transformer architectures work will be covered in the subsequent chapter on this promising technology (Chapter 6). Attention-based models, predominantly Transformers, have been used successfully for some time in sequential data processing, and are now considered the superior alternative to recurrent networks like Long-Short-Term-Memory (LSTM) networks. Transformers have, however, only recently made their way into the image analysis and now also the image generation world. In this section, we will only highlight some developments in the area of generative tasks.

Google Brain/Google AI's 2018 publication on so-called Image Transformers~\cite{parmar2018image}, among other tasks, shows successful conditional image generation for low-resolution 

OpenAI have later shown that even unmodified language Transformers can succeed to model image data, by dealing in sheer compute power for hand-modeling of domain knowledge, which was the basis for the great success of previous unsupervised image generation models. They have trained Image GPT (or iGPT for short), a multi-billion parameter language transformer model, and it excels in several image generation tasks, though only for fairly small image sizes~\cite{pmlr-v119-chen20s}

In the recent past, StyleSwin has been proposed by Microsoft Research Asia~\cite{zhang2021styleswin}, enabling high-resolution image generation. However, the approach uses a block-wise attention window, thereby potentially introducing spatial incoherencies at block edges, which they have to correct for.  

``Taming transformers''~\cite{esser2021taming}, another recent publication already mentioned above, uses what the authors call a learned template code book of image components, which is combined with a Vector Quantization GAN (VQGAN). The VQGAN is structurally modeled after the VQ-VAE, but adds a discriminator network. A transformer model in this architecture composes these code book elements and is interrogated by the GANs variational latent space, conditioned on a textual input, a label image, or other possible inputs. The GAN reconstructs the image from the so-quantized latent space using a combination of a perceptual loss assessing the overall image structure, and a patch-based high-resolution reconstruction loss. By using a sliding attention window approach, the authors prevent patch border artifacts known from StyleSwin. Conditioning on textual input makes use of parts of the CLIP~\cite{radford2021clip-learning} idea (``Contrastive Language-Image Pretraining''), where a language model was train in conjunction with an image encoder to learn embeddings of text-image pairs, sufficient to solve many image understanding tasks with competitive precision, without specific domain adaption. 

It is evidenced by the lineup of institutions that training image transformer models successfully is nothing that can be achieved with modest hardware or on even a medium-scale image database. In particular for the medical area, where data is comparatively scarce even under best assumptions, the power of such models will only be available in the near future if domain transfer learning can be successfully achieved. This, however, is a known strength of transformer architectures. 

\section*{Acknowledgments}
I thank my colleague at the Fraunhofer Institute for Digital Medicine MEVIS, Till Nicke, for his thorough review of the chapter and many valuable suggestions for improvements. I owe many thanks more to other colleagues for their insights both in targeted discussions and most importantly in every-day work life. 

\bibliographystyle{spbasicsort}
\bibliography{references}

\end{document}